\documentclass[]{mosi}

\usepackage{amsmath}
\usepackage{amsfonts}
\usepackage{amssymb}
\usepackage{nicefrac}
\usepackage{natbib}

\usepackage{graphicx}
\usepackage{float}
\usepackage[export]{adjustbox}
\usepackage{booktabs}
\usepackage{array}
\usepackage{multirow}
\usepackage{makecell}
\usepackage{tabularx}
\usepackage{longtable}
\usepackage{threeparttable}
\usepackage[table]{xcolor}
\usepackage{pgfplots}
\pgfplotsset{compat=1.18}
\usetikzlibrary{arrows.meta,positioning,fit}

\usepackage{hyperref}
\usepackage{url}
\usepackage{enumitem}
\usepackage{ragged2e}
\usepackage{pifont}
\usepackage{microtype}
\usepackage{fontspec}
\usepackage{etoolbox}
\usepackage{xparse}

\definecolor{oursgray}{gray}{0.95}
\definecolor{MossCyan}{HTML}{82D9FF}
\definecolor{MossBlue}{HTML}{82B1FF}
\definecolor{tickG}{HTML}{00C853}
\definecolor{crossR}{HTML}{FF1744}
\definecolor{OpenETANavy}{HTML}{123A68}
\definecolor{ToolRowLight}{HTML}{F4F8FC}
\definecolor{ToolRowDark}{HTML}{E7F0FA}

\definecolor{LunaBlue}{HTML}{2F80ED}
\definecolor{TerraGreen}{HTML}{27AE60}
\definecolor{SolYellow}{HTML}{F2C94C}

\newcommand{\tool}[1]{\texttt{\detokenize{#1}}}
\newcommand{\tabletool}[1]{\nolinkurl{#1}}
\newcommand{\toolsep}{,\allowbreak\ }
\newcommand{\faGithub}{%
  \raisebox{-0.2ex}{\includegraphics[height=2.0ex]{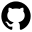}}%
}
\NewDocumentCommand{\codexcomment}{+m}{}

\newcommand{\SelfEvolutionMaxTurns}{100}
\newcommand{\SelfEvolutionMaxToolCalls}{400}
\newcommand{\SelfEvolutionMaxTokens}{10000000}
\newcommand{\SelfEvolutionTimeoutSeconds}{1800}
\newcommand{\SelfEvolutionPlannerIdentityStatus}{not frozen in run manifests}
\newcommand{\SelfStrategyTwoAcceptedCandidates}{2}
\newcommand{\SelfStrategyTwoAuthorModel}{gpt-5.6-luna}
\newcommand{\SelfStrategyTwoAuthorProvider}{AI Gateway}
\newcommand{\SelfStrategyTwoReviewerModel}{gpt-5.6-luna}

\newcommand{\SelfOnlineTaskCount}{10}
\newcommand{\SelfOnlineOneRoundOne}{0}
\newcommand{\SelfOnlineOneRoundTwo}{1}
\newcommand{\SelfOnlineOneRoundThree}{0}
\newcommand{\SelfOnlineTwoRoundOne}{0}
\newcommand{\SelfOnlineTwoRoundTwo}{0}
\newcommand{\SelfOnlineTwoRoundThree}{3}

\newcommand{\SelfTaskLocalSeeds}{10}
\newcommand{\SelfTaskLocalTaskCount}{3}
\newcommand{\SelfTaskLocalEpisodesPerArm}{30}
\newcommand{\SelfTaskLocalEpisodesTotal}{60}
\newcommand{\SelfTaskOneOriginalSuccess}{0}
\newcommand{\SelfTaskOneOriginalAttachment}{2}
\newcommand{\SelfTaskOneOriginalAnyPlace}{1}
\newcommand{\SelfTaskOneCandidateSuccess}{0}
\newcommand{\SelfTaskOneCandidateAttachment}{0}
\newcommand{\SelfTaskOneCandidateAnyPlace}{0}
\newcommand{\SelfTaskTwoOriginalSuccess}{0}
\newcommand{\SelfTaskTwoOriginalAttachment}{3}
\newcommand{\SelfTaskTwoOriginalAnyPlace}{2}
\newcommand{\SelfTaskTwoCandidateSuccess}{0}
\newcommand{\SelfTaskTwoCandidateAttachment}{1}
\newcommand{\SelfTaskTwoCandidateAnyPlace}{0}
\newcommand{\SelfTaskFourOriginalSuccess}{0}
\newcommand{\SelfTaskFourOriginalAttachment}{0}
\newcommand{\SelfTaskFourOriginalAnyPlace}{0}
\newcommand{\SelfTaskFourCandidateSuccess}{0}
\newcommand{\SelfTaskFourCandidateAttachment}{0}
\newcommand{\SelfTaskFourCandidateAnyPlace}{0}
\newcommand{\SelfTaskLocalOriginalSuccess}{0}
\newcommand{\SelfTaskLocalOriginalAttachment}{5}
\newcommand{\SelfTaskLocalOriginalAnyPlace}{3}
\newcommand{\SelfTaskLocalCandidateSuccess}{0}
\newcommand{\SelfTaskLocalCandidateAttachment}{1}
\newcommand{\SelfTaskLocalCandidateAnyPlace}{0}

\newcommand{\SelfPlaybookSeeds}{10}
\newcommand{\SelfPlaybookTaskCount}{3}
\newcommand{\SelfPlaybookEpisodesPerArm}{30}
\newcommand{\SelfPlaybookEpisodesTotal}{60}
\newcommand{\SelfPlaybookTaskOneBaselineSuccess}{2}
\newcommand{\SelfPlaybookTaskOneCandidateSuccess}{0}
\newcommand{\SelfPlaybookTaskTwoBaselineSuccess}{2}
\newcommand{\SelfPlaybookTaskTwoCandidateSuccess}{1}
\newcommand{\SelfPlaybookTaskFourBaselineSuccess}{0}
\newcommand{\SelfPlaybookTaskFourCandidateSuccess}{0}
\newcommand{\SelfPlaybookBaselineSuccess}{4}
\newcommand{\SelfPlaybookBaselineRatePercent}{13.3}
\newcommand{\SelfPlaybookCandidateSuccess}{1}
\newcommand{\SelfPlaybookCandidateRatePercent}{3.3}

\newcommand{\SelfPlaybookBaselineOnly}{4}
\newcommand{\SelfPlaybookCandidateOnly}{1}
\newcommand{\SelfPlaybookBothSuccess}{0}
\newcommand{\SelfPlaybookBothFail}{25}
\newcommand{\SelfPlaybookMcNemarP}{0.375}

\newcommand{\SelfPlaybookBaselineGraspEstimate}{30}
\newcommand{\SelfPlaybookBaselineMove}{27}
\newcommand{\SelfPlaybookBaselineClose}{21}
\newcommand{\SelfPlaybookBaselineAttachmentAssess}{16}
\newcommand{\SelfPlaybookBaselineAttachmentPass}{14}
\newcommand{\SelfPlaybookBaselineAnyPlace}{12}
\newcommand{\SelfPlaybookBaselineRelease}{15}
\newcommand{\SelfPlaybookBaselineReward}{4}
\newcommand{\SelfPlaybookCandidateGraspEstimate}{30}
\newcommand{\SelfPlaybookCandidateMove}{24}
\newcommand{\SelfPlaybookCandidateClose}{15}
\newcommand{\SelfPlaybookCandidateAttachmentAssess}{5}
\newcommand{\SelfPlaybookCandidateAttachmentPass}{5}
\newcommand{\SelfPlaybookCandidateAnyPlace}{4}
\newcommand{\SelfPlaybookCandidateRelease}{10}
\newcommand{\SelfPlaybookCandidateReward}{1}
\newcommand{\SelfPlaybookBaselineMeanCalls}{42.6}
\newcommand{\SelfPlaybookBaselineMeanTurns}{42.8}
\newcommand{\SelfPlaybookBaselineMeanSeconds}{1546.2}
\newcommand{\SelfPlaybookBaselineTimeouts}{18}
\newcommand{\SelfPlaybookCandidateMeanCalls}{49.2}
\newcommand{\SelfPlaybookCandidateMeanTurns}{49.3}
\newcommand{\SelfPlaybookCandidateMeanSeconds}{1678.2}
\newcommand{\SelfPlaybookCandidateTimeouts}{23}
\newcommand{\SelfPlaybookCaseTask}{1}
\newcommand{\SelfPlaybookCaseSeed}{3}
\newcommand{\SelfPlaybookCaseBaselineSuccess}{1}
\newcommand{\SelfPlaybookCaseBaselineTurns}{26}
\newcommand{\SelfPlaybookCaseBaselineCalls}{26}
\newcommand{\SelfPlaybookCaseBaselineSeconds}{1093.8}
\newcommand{\SelfPlaybookCaseCandidateSuccess}{0}
\newcommand{\SelfPlaybookCaseCandidateTurns}{51}
\newcommand{\SelfPlaybookCaseCandidateCalls}{51}
\newcommand{\SelfPlaybookCaseCandidateSeconds}{1801.0}

\newcommand{\SelfStrategyOnePairCount}{3}

\newcommand{\SelfStrategyOneBaselineSuccess}{0}
\newcommand{\SelfStrategyOneBaselineMeanTurns}{43.3}
\newcommand{\SelfStrategyOneBaselineMeanSeconds}{1342.3}
\newcommand{\SelfStrategyOneCandidateSuccess}{0}
\newcommand{\SelfStrategyOneCandidateMeanTurns}{46.7}
\newcommand{\SelfStrategyOneCandidateMeanSeconds}{1634.9}
\newcommand{\SelfStrategyTwoPairCount}{1}
\newcommand{\SelfStrategyTwoBaselineSuccess}{0}
\newcommand{\SelfStrategyTwoBaselineTurns}{42}
\newcommand{\SelfStrategyTwoBaselineCalls}{42}
\newcommand{\SelfStrategyTwoBaselineTokens}{777316}
\newcommand{\SelfStrategyTwoBaselineSeconds}{1801.1}
\newcommand{\SelfStrategyTwoBaselineViolations}{0}
\newcommand{\SelfStrategyTwoCandidateSuccess}{0}
\newcommand{\SelfStrategyTwoCandidateTurns}{34}
\newcommand{\SelfStrategyTwoCandidateCalls}{33}
\newcommand{\SelfStrategyTwoCandidateTokens}{973525}
\newcommand{\SelfStrategyTwoCandidateSeconds}{980.2}
\newcommand{\SelfStrategyTwoCandidateViolations}{1}
\newcommand{\SelfStrategyTwoTask}{2}
\newcommand{\SelfStrategyTwoSeed}{0}
\newcommand{\SelfStrategyTwoExcludedInfrastructure}{0}

\newcommand{\SelfOnlineEvidenceOne}{spatial-adaptive-3round-20260728-r1}
\newcommand{\SelfOnlineEvidenceTwo}{spatial-adaptive-3round-20260729-r2}
\newcommand{\SelfTaskLocalEvidence}{spatial-self-evolution-ab-20260729-r1}
\newcommand{\SelfPlaybookEvidence}{spatial-playbook-ab-20260729-r1}
\newcommand{\SelfStrategyOneEvidence}{spatial-task-strategy-replay-20260729-r3}
\newcommand{\SelfStrategyTwoEvidence}{spatial-contrastive-strategy-v2-20260729-r3}

\newcommand{\LiberoResultStatus}{frozen}

\newcommand{\LiberoProtocolKind}{fixed-matrix}
\newcommand{\LiberoSuiteCount}{4}
\newcommand{\LiberoTasksPerSuite}{10}
\newcommand{\LiberoTaskCount}{40}
\newcommand{\LiberoSeedCount}{10}
\newcommand{\LiberoEpisodeSuccesses}{56}
\newcommand{\LiberoEpisodeDenominator}{400}
\newcommand{\LiberoTasksWithSuccess}{18}
\newcommand{\LiberoInfrastructureInvalid}{3}

\newcommand{\LiberoDiagnosticDenominator}{397}

\newcommand{\LiberoResourceEpisodeCount}{397}

\newcommand{\MinimalTaskCount}{130}

\newcommand{\MinimalLunaPassOneSuccesses}{21}

\newcommand{\MinimalLunaPassFiveSuccesses}{62}
\newcommand{\MinimalLunaPassFivePercent}{47.7}

\newcommand{\MinimalTerraPassOneSuccesses}{58}

\newcommand{\MinimalTerraPassFiveSuccesses}{83}
\newcommand{\MinimalTerraPassFivePercent}{63.8}

\newcommand{\MinimalSolPassOneSuccesses}{92}

\newcommand{\MinimalSolPassFiveSuccesses}{117}
\newcommand{\MinimalSolPassFivePercent}{90.0}

\newcommand{\RealRobotResultStatus}{pending}

\title{ETA: A New Agentic Paradigm for Embodied Tasks}

\author{
Yitong Chen$^{1,2,*}$, Zezheng Huai$^{1,3,*}$, Sixian Li$^{1,2,*}$, Yubang Wang$^{1,2,*}$, Haozhe Zhang$^{1,5,*}$\\ 
Yifei Zhang$^{1,4,*}$, Hechang Chen$^{1,3,\dagger}$
 Jingjing Gong$^{1,\dagger}$, Yu-Gang Jiang$^{2,\dagger}$,Xipeng Qiu$^{1,2,\dagger}$
\\[2mm]
{\normalfont\normalsize
\mbox{$^{1}$Shanghai Innovation Institute}\qquad
\mbox{$^{2}$Fudan University}\qquad
\mbox{$^{3}$Jilin University}\\[1mm]
\mbox{$^{4}$Nanjing University}\qquad
\mbox{$^{5}$Zhejiang University}
}
}

\abstract{
When will robots have their ChatGPT moment? Such a breakthrough requires a
general-purpose robot that can handle unfamiliar tasks in unfamiliar
environments, remain controllable over long interactions, and learn from
experience.

Today's embodied systems largely follow an end-to-end observation-to-action
path. Despite rapid progress, they remain far from this goal: their
generalization depends heavily on the coverage of robot training data, while
long task execution remains difficult to control and inspect. To realize this
goal, we introduce the \emph{Embodied Task Agent} (ETA), a new paradigm for
extending digital agents into the physical world, and release OpenETA as its
open-source implementation. ETA centers the robot around a Planner that
chooses one Tool call at a time, an Interface that controls execution, and a
World that returns the result and a fresh observation. This loop allows the
agent to verify outcomes, adapt its plan, and turn successful and failed
interactions into reusable experience. OpenETA provides replaceable Planners,
composable Tools and Skills, auditable memory, replayable trajectories, and
common interfaces for simulation and real robots. For Codex, OpenETA can
operate as a lightweight plugin that exposes only \tool{observe},
\tool{mark_point}, and \tool{move_to}.

Without using a VLA or task-specific policy as a Tool, OpenETA with
gpt-5.6-Sol reaches \textsc{Pass}@5 of
\MinimalSolPassFiveSuccesses/\MinimalTaskCount{} (\MinimalSolPassFivePercent\%)
on 130 LIBERO tasks. Sol already solves
\MinimalSolPassOneSuccesses/\MinimalTaskCount{} tasks on the first seed.
These results offer a practical step toward general, controllable, and
self-improving physical agents.
\ifdefstring{\LiberoResultStatus}{frozen}
  {}
  {For the formal full-system batch, this draft reports no obsolete aggregate values;
  development trajectories are status evidence, not capability
  evidence.}

\par\vspace{0.9em}
\begin{center}
  \begin{minipage}{0.88\linewidth}
    \centering
    {\fontsize{9.3}{12}\selectfont\itshape\color{OpenETANavy}
    ``We are on the eve of the largest robotics distribution in history.
    By Saturday, there'll be one robot to every five humans.''\par}
    \vspace{0.4em}
    {\raggedleft\fontsize{8.2}{10}\selectfont\normalfont\color{OpenETANavy}
    --- \textit{I, Robot} (2004)\par}
  \end{minipage}
\end{center}
}

\checkdata[\raisebox{-0.1ex}{\faGithub}\hspace{0.4em}GitHub]{%
  \url{https://github.com/OpenMOSS/OpenETA}%
}
\checkdata[Project Page]{%
  \url{https://openmoss.ai/OpenETA/}%
}

\begin{document}
\maketitle
\begingroup
\renewcommand{\thefootnote}{\fnsymbol{footnote}}
\footnotetext[1]{%
  Equal contribution, sort by the first letter of the surname.
  $^{\dagger}$Corresponding authors.
  Emails:
  \href{mailto:cyt050719@mail.ustc.edu.cn}{cyt050719@mail.ustc.edu.cn},
  \href{mailto:253208540294@sii.edu.cn}{253208540294@sii.edu.cn},
  \href{mailto:253308120308@sii.edu.cn}{253308120308@sii.edu.cn},
  \href{mailto:ybangwang518@whu.edu.cn}{ybangwang518@whu.edu.cn},
  \href{mailto:zhanghaozhe@sii.edu.cn}{zhanghaozhe@sii.edu.cn}, and
  \href{mailto:yifeizhang@sii.edu.cn}{yifeizhang@sii.edu.cn}.%
}
\endgroup

\section{Introduction}
\label{sec:introduction}

Recent large language models have moved beyond text generation, showing broad
abilities in reasoning, planning, tool use, and error correction. Coding agents
make this emergence concrete. They can decompose high-level goals, navigate
complex digital environments, execute actions, evaluate results, recover from
failures, and preserve successful procedures as reusable experience
\citep{DBLP:conf/iclr/YaoZYDSN023,DBLP:conf/nips/ShinnCGNY23,
DBLP:journals/tmlr/WangX0MXZFA24}. Coding agents are now being applied across
software engineering, data analysis, scientific research, automated
experimentation, and model design and optimization, with a growing ability to
conduct long workflows and improve their own strategies through execution
feedback \citep{karpathy2026autoresearch,zhang2026hyperagents}.
Together, these developments point toward a general digital agent that solves
diverse tasks through execution, feedback, and accumulated experience.

This emergence suggests a practical picture of embodied intelligence's ChatGPT
moment. We define such a system through three core capabilities. First,
\emph{generality}: the same system should understand new tasks and reuse or
compose perception, planning, and control capabilities across objects,
environments, and embodiments, rather than require a separate policy for every
task. Second, \emph{controllability}: it should maintain explicit task state
over long horizons, expose each physical action through a bounded interface,
and use actual execution results to determine what happens next. Third,
\emph{self-improvement}: it should turn successful and failed interactions into
reusable memory and Skills, while validating each update before it changes
future behavior. These are properties of the whole agent system, in which
models provide capabilities and the agent composes, monitors, and improves
them over time.

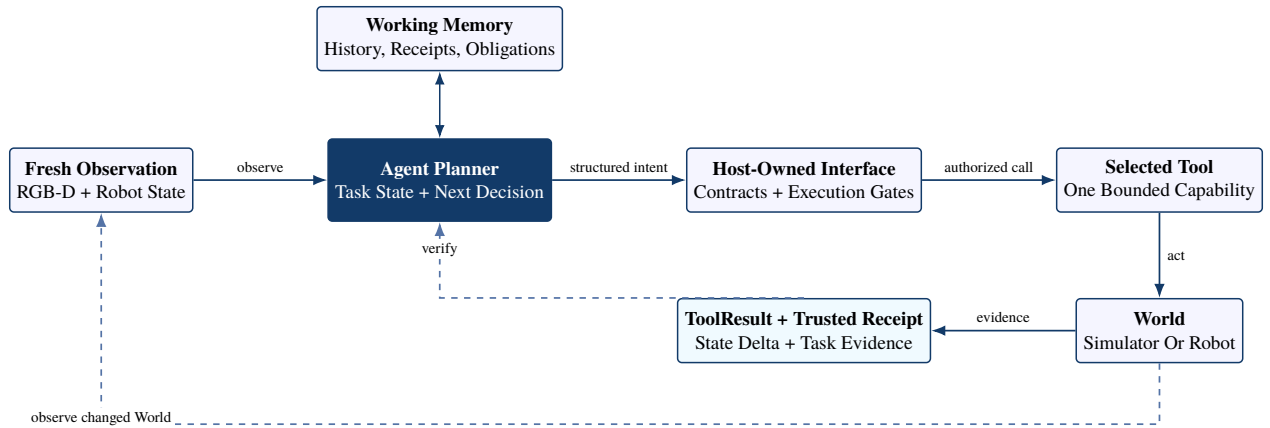
\begin{figure}[htbp]
  \centering
  \resizebox{\textwidth}{!}{%
  \begin{tikzpicture}[
    >=Latex,
    node distance=11mm and 22mm,
    every node/.style={font=\small,align=center},
    role/.style={draw=OpenETANavy,rounded corners=2pt,fill=blue!4,
      minimum width=2.55cm,minimum height=1.05cm,line width=0.7pt},
    planner/.style={role,fill=OpenETANavy,text=white,minimum width=3.15cm,
      minimum height=1.35cm,font=\small},
    evidence/.style={role,fill=MossCyan!12},
    flow/.style={->,line width=0.8pt,draw=OpenETANavy},
    feedback/.style={->,line width=0.8pt,draw=MossBlue!65!black,dashed}
  ]
    \node[planner] (planner) {\textbf{Agent Planner}\\Task State + Next Decision};
    \node[role,above=of planner] (memory) {\textbf{Working Memory}\\History, Receipts, Obligations};
    \node[role,left=of planner] (observation) {\textbf{Fresh Observation}\\RGB-D + Robot State};
    \node[role,right=of planner] (interface) {\textbf{Host-Owned Interface}\\Contracts + Execution Gates};
    \node[role,right=of interface] (tool) {\textbf{Selected Tool}\\One Bounded Capability};
    \node[role,below=14mm of tool] (world) {\textbf{World}\\Simulator Or Robot};
    
    \node[evidence,below=14mm of interface] (receipt) {\textbf{ToolResult + Trusted Receipt}\\State Delta + Task Evidence};

    \draw[flow] (observation) -- node[above,font=\scriptsize]{observe} (planner);
    \draw[flow] (planner) -- node[above,font=\scriptsize]{structured intent} (interface);
    \draw[flow] (interface) -- node[above,font=\scriptsize]{authorized call} (tool);
    \draw[flow] (tool) -- node[right,font=\scriptsize]{act} (world);
    \draw[flow] (world) -- node[above,font=\scriptsize]{evidence} (receipt);
    \draw[feedback] (receipt.north) -|
      node[pos=0.72,above=2pt,fill=white,inner sep=1.5pt,font=\scriptsize]{verify}
      (planner.south);
    \draw[feedback] (world.south) -- ++(0,-10mm) -|
      node[pos=0.55,below=2pt,fill=white,inner sep=1.5pt,font=\scriptsize]
      {observe changed World} (observation.south);
    \draw[<->,line width=0.7pt,draw=OpenETANavy] (memory) -- (planner);
  \end{tikzpicture}%
  }
  \caption{\textbf{Planner-centered information flow in OpenETA.}
  The Planner chooses each task-level step, the Interface checks and executes
  the Tool call, and the World returns evidence and a fresh observation before
  the next decision.}
  \label{fig:openeta-overview}
\end{figure}

However, embodied intelligence today still focuses mainly on end-to-end
observation-to-action learning. Vision--language--action models (VLAs) map
visual observations and instructions to robot actions
\citep{DBLP:conf/corl/ZitkovichYXXXXW23,DBLP:conf/corl/KimPKXB0RFSVKBT24},
while world action models (WAMs) also predict possible future states and
actions \citep{DBLP:journals/corr/abs-2605-12090}. Despite rapid progress,
their generalization depends heavily on the coverage of robot training data,
and behavior over long tasks remains difficult to control and inspect
\citep{DBLP:journals/corr/abs-2510-03827}. To move toward the vision above, we
introduce the \emph{Embodied Task Agent} (ETA), a new paradigm for extending
digital agents into the physical world, and release OpenETA as its open-source
implementation. ETA adds a task-level Planner above specialized models and
Tools. As Figure~\ref{fig:openeta-overview} shows, the Planner chooses one Tool
call, the Interface controls its execution, and the World returns the result
and a fresh observation before the next decision. OpenETA makes this design
open and extensible through composable Tools and Skills, replaceable Planners,
simulation and robot backends, auditable memory, and replayable trajectories.
OpenETA can also operate as a lightweight plugin for Codex that uses only
\tool{observe}, \tool{mark_point}, and \tool{move_to}.

\paragraph{Contributions}
\begin{itemize}[leftmargin=*]
  \item \textbf{A new agentic paradigm for embodied tasks.} ETA places a Planner
  at the center of the physical task loop. The Planner can issue only one
  world-changing Tool call before it must receive a fresh observation and
  decide again.
  \item \textbf{An open-source framework.} OpenETA implements ETA with modular
  Planners, Tools, memory, and simulation-to-robot interfaces. Its execution
  records make each run easy to inspect and replay. We release OpenETA with
  both a broad Tool registry and a lightweight three-Tool plugin for Codex.
  \item \textbf{Simulation results and deployment resources.} Without using a
  VLA or task-specific policy as a Tool, OpenETA is evaluated with gpt-5.6-Luna,
  gpt-5.6-Terra, and gpt-5.6-Sol with medium reasoning effort. At
  \textsc{Pass}@5, they solve
  \MinimalLunaPassFiveSuccesses/\MinimalTaskCount{} (\MinimalLunaPassFivePercent\%),
  \MinimalTerraPassFiveSuccesses/\MinimalTaskCount{} (\MinimalTerraPassFivePercent\%),
  and \MinimalSolPassFiveSuccesses/\MinimalTaskCount{} (\MinimalSolPassFivePercent\%)
  LIBERO tasks, respectively. We also release experiment scripts and an interface-level real-robot integration
  for a UR5e--Robotiq platform.
\end{itemize}

The rest of the paper reviews related work, defines ETA, and describes OpenETA.
It then reports simulation studies, constrained self-evolution experiments,
and physical deployment, followed by limitations and future work. The
appendices provide a detailed pick-and-place trace and additional
reproducibility material.

\section{Related Work}
\label{sec:related}

\subsection{From language models to coding agents}

Language-model agents extend generation into an iterative interaction loop.
ReAct interleaves reasoning with actions and observations
\citep{DBLP:conf/iclr/YaoZYDSN023}. Modern coding agents apply the same idea
through stable interfaces for files, shells, search, and version control. The
model selects an operation, the runtime executes it, and the next decision uses
the updated external state. Recent systems also expose robot operation through
general-purpose agent interfaces, as illustrated by Anthropic's robotics
demonstration \citep{berman2026claude}.

A related line improves agents through non-parametric experience. Reflexion
stores verbal reflections \citep{DBLP:conf/nips/ShinnCGNY23}, ExpeL extracts
reusable experience from training tasks \citep{DBLP:conf/aaai/Zhao0XLLH24},
and Voyager grows a verified code-skill library
\citep{DBLP:journals/tmlr/WangX0MXZFA24}. These methods show how an Agent can
reuse experience, which motivates OpenETA's memory and Skill layers. In a
physical system, however, a bad memory or Skill can lead to a bad action.
OpenETA therefore lets new experience affect later execution only after it
reproduces task success and passes a paired safety check
(Section~\ref{sec:self_revolution}).

\subsection{Embodied foundation models: VLAs and world models}

Embodied foundation models combine large multimodal models with robot data.
PaLM-E established embodied multimodal language modeling
\citep{DBLP:conf/icml/DriessXSLCIWTVY23}, and Open X-Embodiment brought together
heterogeneous robot datasets and RT-X models
\citep{DBLP:journals/corr/abs-2310-08864}. Vision--language--action models
(VLAs) go further by mapping observations and language to robot actions. This
line includes RT-2 and OpenVLA
\citep{DBLP:conf/corl/ZitkovichYXXXXW23,DBLP:conf/corl/KimPKXB0RFSVKBT24},
Octo and $\pi_0$
\citep{DBLP:conf/rss/GhoshWPBMDHK0LT24,DBLP:journals/corr/abs-2410-24164}, and
GR00T N1 for humanoid robots \citep{DBLP:journals/corr/abs-2503-14734}.
FAST studies more efficient action tokenization for this model family
\citep{DBLP:journals/corr/abs-2501-09747}.
Recent systems extend this direction to unseen homes with $\pi_{0.5}$
\citep{DBLP:journals/corr/abs-2504-16054} and to contact-rich or deformable
objects with CoRE-VLA
\citep{zhang2026corevlascalablerobustvisionlanguageaction}. OpenETA treats
these models as specialist Tools. The Planner can invoke them when a task is
hard to express with geometric Tools, while the Interface still checks the
action and verifies its result.

\begin{figure}[htbp]
  \centering
  \includegraphics[width=\textwidth]{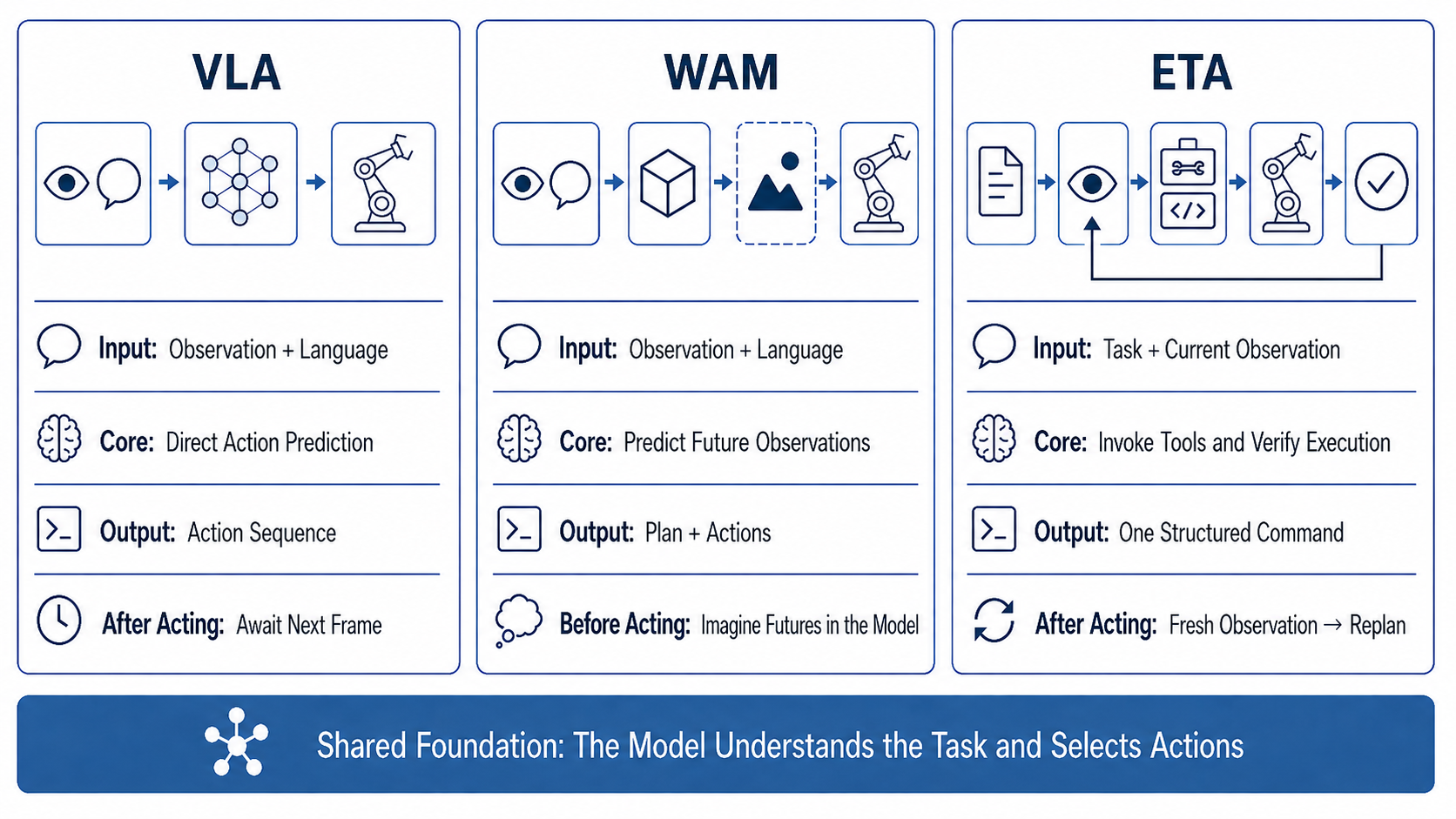}
  \caption{\textbf{From action prediction to task-level closed loops: different
  roles of VLA, WAM, and ETA.} All three paradigms share a foundation model
  that understands the task and selects actions. A VLA directly predicts an
  action sequence, while a WAM imagines possible future observations and
  actions before execution. An ETA instead issues structured commands through
  Tools and incorporates fresh post-action World evidence into a task-level
  runtime closed loop.}
  \label{fig:vla-wam-eta}
\end{figure}

World models add explicit prediction of physical futures. DreamZero jointly
predicts video and action for closed-loop control and cross-embodiment transfer
\citep{DBLP:journals/corr/abs-2602-15922}; DreamDojo pretrains a generalist robot world model on
large-scale egocentric human video and supports prediction, planning, and
policy evaluation \citep{DBLP:journals/corr/abs-2602-06949}. More broadly, WAMs organize future
observation and action prediction in one foundation-model paradigm
\citep{DBLP:journals/corr/abs-2605-12090}. Predicted futures can rank plans or provide subgoals, but
they are not observations of the deployed World. ETA therefore treats a VLA
as an action Tool and a world model as a prediction or plan-comparison Tool;
trusted physical evidence still determines the next task-level decision.
Figure~\ref{fig:vla-wam-eta} summarizes this division of labor: a VLA proposes
actions, a WAM predicts possible futures, and an ETA coordinates heterogeneous
capabilities around observed World evidence.

\subsection{VLM coding models for robot tasks}

Early systems connected language models to predefined robot Skills and APIs.
SayCan ranks Skills by language relevance and learned affordances
\citep{DBLP:conf/corl/IchterBCFHHHIIJ22}. ProgPrompt and Code as Policies
generate executable robot programs
\citep{DBLP:conf/icra/SinghBMGXTFTG23,DBLP:conf/icra/LiangHXXHIFZ23}, while
Inner Monologue returns scene descriptions and success signals to
language-level planning \citep{DBLP:conf/corl/HuangXXCLFZTMCS22}. ChatGPT for
Robotics describes general design principles for connecting language models to
robot APIs \citep{DBLP:journals/access/VempralaBBK24}.

Later work gives these programs more explicit spatial structure. Instruct2Act
combines vision models with robot primitives
\citep{DBLP:journals/corr/abs-2305-11176}. Text2Motion and SayPlan add
feasibility checks or 3D scene graphs
\citep{DBLP:journals/arobots/LinAMPB23,DBLP:journals/corr/abs-2307-06135},
while VoxPoser and RoboTool generate spatial constraints or structured code
\citep{DBLP:conf/corl/HuangWZL0023,DBLP:journals/corr/abs-2310-13065}.
MOKA, CoPa, ReKep, and OmniManip represent manipulation through keypoints,
waypoints, costs, or interaction geometry
\citep{DBLP:conf/rss/FangLAL24,DBLP:conf/iros/HuangLHW024,
DBLP:conf/corl/HuangWLZF24,DBLP:conf/cvpr/PanZWZGD25}.
RoboScript and RoboCodeX pursue reusable and object-centric robot code
\citep{DBLP:journals/corr/abs-2402-14623,
DBLP:conf/icml/0001CZCYGCLHTSY24}. OK-Robot integrates open components, while
Manipulate-Anything reduces reliance on privileged state
\citep{DBLP:journals/corr/abs-2401-12202,
DBLP:conf/corl/DuanYPWEFK24}. Reflective Planning evaluates imagined future
states before choosing a Skill \citep{DBLP:journals/corr/abs-2502-16707}.

Recent systems add richer execution feedback. CaP-X revises programs from
single- or multi-turn feedback and can synthesize new Skills
\citep{DBLP:journals/corr/abs-2603-22435}. ASPIRE uses multimodal execution
traces to locate failures and validate program repairs
\citep{lu2026aspireagenticskillsdiscovery}. VIA uses a visual interface as the
Agent's robot-control surface \citep{hu2026viavisualinterfaceagent}, and
omnimodal embodied agents combine robot models, smart-home services, Web
search, interaction, and memory in one task loop
\citep{DBLP:journals/corr/abs-2606-27251}.

These systems are not simply open loop: a generated program can query the
World and contain reactive control. The main difference is when the VLM makes
the next high-level decision. Program-centric methods generate a composition
and then repair or refine it around execution. OpenETA instead selects and
combines registered capabilities during execution. After every
world-changing Tool call, the Planner receives a fresh observation before it
chooses the next Tool. This design makes each physical step easy to trace and
control, but it requires more Planner calls and increases inference latency.

\section{The Embodied Task Agent Paradigm}
\label{sec:eta}

ETA does not give a model unrestricted control of the world. Instead, it
exposes bounded, verifiable atomic capabilities. An ETA is a task-level Agent,
not another next-action predictor. It interprets a natural-language goal,
decomposes the task, and selects Tools, Skills, and AtomActions. After each
action, it reads the new observation and environment receipt. It then decides
whether to continue, retry, replan, or request help.

\begin{figure}[htbp]
  \centering
  \includegraphics[width=\textwidth,trim=0 105 0 145,clip]{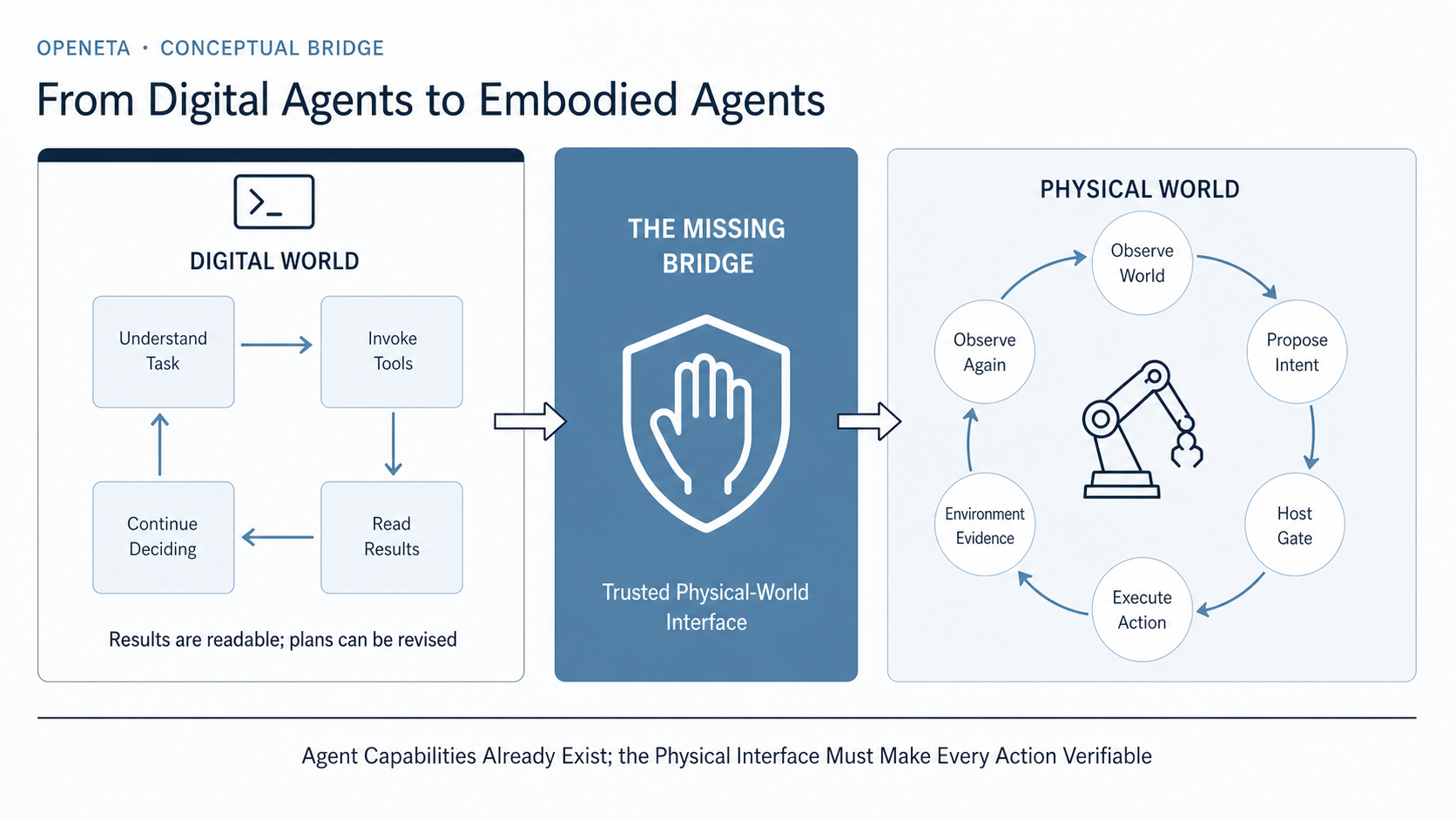}
  \caption{\textbf{From digital agents to embodied agents.}
  Digital Tool results are directly readable; a physical Agent additionally
  needs host-gated execution, fresh observations, and trusted evidence of what
  changed in the World.}
  \label{fig:digital-embodied}
\end{figure}

Figure~\ref{fig:digital-embodied} summarizes the conceptual bridge. Digital
agents already alternate between understanding a task, invoking Tools, and
reading results. ETA keeps this loop but routes every command that can change
the physical World through the Interface. The Interface checks the command and
authorizes its execution; the Agent cannot act on the World directly.

At turn \(t\), the ETA produces a structured command \(c_t\) from the current
observation \(o_t\), goal \(g\), and working memory \(m_t\):
\begin{equation}
  c_t = \pi_{\mathrm{agent}}(g, o_t, m_t).
\end{equation}
The Interface validates the command structure, Tool contract, provenance,
authority, and prerequisite evidence before dispatching an atomic action
\(a_t\):
\begin{equation}
  a_t =
  \begin{cases}
    \mathrm{dispatch}(c_t), & \mathrm{gate}(c_t, o_t, m_t)=\mathrm{pass},\\
    \varnothing, & \text{otherwise}.
  \end{cases}
\end{equation}
The World returns a Tool result \(y_t\), an environment receipt \(e_t\), and a
fresh observation \(o_{t+1}\). Planning resumes only after these records enter
working memory.

\subsection{Runtime invariant}

Every ETA execution obeys:
\begin{quote}
\textbf{Execute only one world-changing action at a time. Afterward, obtain a
fresh observation before executing the next state-dependent action.}
\end{quote}

This rule prevents an unobserved sequence of world changes and prevents the
system from assuming that the scene remained unchanged. If an action succeeds,
or its transport status is unknown, without a sufficiently fresh state
snapshot, the Interface creates a fresh-observation obligation. The system
must refresh state first; after repeated observation failures, it stops safely
instead of guessing from stale images.

\subsection{Three roles and their trust boundary}

ETA separates the system into:
\begin{enumerate}[leftmargin=*]
  \item \textbf{Agent}: understands the task, maintains working memory, and
  proposes the next structured command.
  \item \textbf{Interface}: validates structure, authority, provenance, and
  prerequisites, and owns Tool dispatch and execution gates.
  \item \textbf{World}: executes actions. A simulator can expose ground-truth
  state, contacts, reward, and termination. A physical system directly exposes
  robot state, while task relations and completion often require sensors,
  checker models, or human judgment.
\end{enumerate}

\FloatBarrier

\section{The OpenETA System}
\label{sec:openeta}

OpenETA is an open-source runtime and experimental framework for the ETA
paradigm. It composes foundation models, Tools, Skills, an Interface, World
backends, and evaluation in one observe--decide--act--verify loop. It does not
require a monolithic model to predict every robot action. We provide two
configurations. Full OpenETA uses a broad Tool registry, while OpenETA for
Codex is a lightweight plugin that exposes only three embodied Tools. Both
configurations follow the same
Agent--Interface--World protocol; Section~\ref{sec:evaluation} gives their
benchmark settings.

We use \emph{host} to mean this runtime: the software that registers Tools,
checks Agent commands, and dispatches approved actions. The host is outside the
model-controlled Agent.

OpenETA separates intelligence from execution authority: the Agent decides
what it wants to do, the Interface decides whether it may, and the World states
what actually happened. Every observation, command, action, receipt, and
decision rationale enters a replayable trajectory, allowing models, Tools,
Skills, World backends, and checkers to be replaced under one protocol.

\subsection{Agent}

\paragraph{Tool}
A Tool is a host-registered atomic capability with stable parameter and return
contracts. The current default registry contains 44 Tools. Each Tool also
declares a side-effect class---\tool{read_only}, \tool{planning},
\tool{bookkeeping}, or \tool{world_mutating}---which determines whether calls
may be batched and whether a fresh observation is required afterward.

\begin{table}[htbp]
  \centering
  \small
  \renewcommand{\arraystretch}{1.28}
  \setlength{\tabcolsep}{6pt}
  \caption{\textbf{Major OpenETA Tool categories.}}
  \label{tab:tools}
  \rowcolors{2}{ToolRowLight}{ToolRowDark}
  \begin{tabularx}{\textwidth}{
    >{\raggedright\arraybackslash}p{0.18\textwidth}
    >{\raggedright\arraybackslash}p{0.36\textwidth}
    >{\raggedright\arraybackslash}X}
    \rowcolor{OpenETANavy}
    \textcolor{white}{\textbf{Category}} &
    \textcolor{white}{\textbf{Representative Tools}} &
    \textcolor{white}{\textbf{Role}} \\
    \textbf{Perception and localization} &
    \tabletool{observe}\toolsep \tabletool{retrieve_asset_reference}\toolsep
    \tabletool{molmopoint}\toolsep \tabletool{sam3}\toolsep \tabletool{enhance_depth} &
    Acquire observations, retrieve target references, localize and segment
    targets, and create traceable visual artifacts. \\
    \textbf{Geometry and manipulation planning} &
    \tabletool{grasp_pose_estimate}\toolsep \tabletool{anygrasp}\toolsep
    \tabletool{contact_graspnet}\toolsep \tabletool{graspgenx}\toolsep \tabletool{anyplace} &
    Generate grasp or placement proposals, transform coordinates, and retain
    candidate provenance and scores. \\
    \textbf{Safety and physical execution} &
    \tabletool{ik_preview_check}\toolsep \tabletool{obstacle_avoidance}\toolsep
    \tabletool{move_to}\toolsep \tabletool{follow_eef_trajectory}\toolsep
    \tabletool{gripper_control} &
    Check feasibility and change robot or environment state through atomic
    actions. \\
    \textbf{Environment and evidence} &
    \tabletool{create_simulator_env}\toolsep \tabletool{close_simulator_env}\toolsep
    \tabletool{materialize_mcp_images} &
    Manage environment lifecycle, materialize remote observations, and attach
    environment receipts. \\
    \textbf{Agent support} &
    \tabletool{save_memory}\toolsep \tabletool{get_memory}\toolsep \tabletool{compact_memory}\toolsep
    \tabletool{python_exec}\toolsep \tabletool{web_search}\toolsep \tabletool{register_skill} &
    Manage working memory, bounded code, information retrieval, and editable
    Skills. \\
  \end{tabularx}
\end{table}

The registry wraps specialist methods behind stable OpenETA contracts rather
than exposing their native APIs directly. Representative backends include SAM
3 for concept-conditioned segmentation
\citep{DBLP:journals/corr/abs-2511-16719}, AnyGrasp and Contact-GraspNet for
6-DoF grasp proposals
\citep{DBLP:journals/trob/FangWFGLYLXL23,DBLP:conf/icra/SundermeyerMTF21}, and
AnyPlace for object placement \citep{DBLP:journals/corr/abs-2502-04531}.
Collision-aware trajectory generation can likewise be supplied by the cuRobo
family \citep{DBLP:journals/corr/abs-2310-17274,DBLP:journals/corr/abs-2603-05493}.
The Interface normalizes these heterogeneous outputs and retains their
provenance, while the Planner still owns the next task-level decision.

\paragraph{Skill}
A Skill is editable textual guidance describing how the model should reason,
check, and recover. It never executes a hidden action sequence; the Agent must
still select every atomic Tool explicitly.

\paragraph{Response and Soul}
Response supports dialogue, help requests, and completion reports. Soul
describes identity, behavioral boundaries, and risk preferences that persist
across tasks and sessions. Both influence decisions but receive no physical
execution authority.

\paragraph{AtomAction}
An AtomAction is an execution primitive that can physically change the World,
such as end-effector motion, trajectory following, or gripper control. It is
the physical subset of \tool{world_mutating} Tools; Skill text cannot bypass
AtomActions to modify the World.

\subsection{Interface}

\paragraph{Stable contracts and normalized results}
The Planner submits only two structured command types: \tool{tool_call} and
\tool{response}. The host runtime defines Tool names, parameters, handlers, and
side-effect declarations. The Agent cannot rewrite them at runtime. Every
return is normalized into \tool{ToolResult}, which separately records outputs,
artifact references, state deltas, diagnostics, and environment receipts.

\paragraph{Execution gates and obligations}
Read-only or planning calls may be batched under bounded conditions, whereas
\tool{world_mutating} actions execute one at a time. The Interface blocks
execution if the target is unconfirmed, a safety check fails, or prerequisite
evidence is missing. Explicit mask selection, grasp-candidate switching, and
post-action observation become obligations in working memory; unresolved
obligations constrain subsequent Tool calls.

\paragraph{Trusted receipts and backend isolation}
The Agent sees stable Tool semantics; the Interface connects them to simulators
or robots through adapter boundaries such as MCP. Low-level joint vectors,
controller expansion, and internal environment objects remain hidden from the
planner. Reward and termination are accepted only when host-attested
provenance, execution ID, and session ID match the current turn; an ordinary
Tool handler cannot mint official reward. Appendix~\ref{app:core-protocol}
specifies the minimum command, result, receipt, obligation, and rollout
schemas.

\subsection{World}

The World layer is implemented by two families of Model Context Protocol (MCP)
services. \textbf{Simulator MCP} owns environment lifecycle, normalized
observations, atomic motion and gripper execution, reward, and termination.
\textbf{Real-robot MCP} exposes the corresponding camera, robot-state, motion,
and gripper capabilities through device-specific drivers and safety limits.
The Interface maps both services to the same planner-facing Tool contracts, so
the Agent does not need simulator- or robot-specific control code.

This separation is also a security and evaluation boundary. The Agent cannot
directly import a simulator object, read privileged ground-truth state, call an
unregistered controller, or fabricate reward. It can access only fields and
operations intentionally exposed by the MCP service. Simulator contacts,
object poses, and official task verdicts may be used internally to construct a
host-attested receipt, but privileged values that are not part of the declared
observation contract never enter Planner context. The same boundary prevents a
powerful coding model from ``solving'' a benchmark by inspecting hidden state
or bypassing the physical action path.

\FloatBarrier

\section{Simulation Experiments}
\label{sec:evaluation}

\subsection{Experimental setup}

\subsubsection{Full OpenETA configuration}

We evaluate OpenETA's complete closed loop on the LIBERO manipulation benchmark
\citep{DBLP:conf/nips/LiuZGFLZS23}. The formal scope comprises Spatial, Object, Goal, and
Long / LIBERO-10: \LiberoSuiteCount{} suites with
\LiberoTasksPerSuite{} tasks each. OpenETA receives no additional
task-specific policy training; it composes foundation-model reasoning, target
localization, segmentation, grasp planning, motion control, and environment
checking Tools. Its Planner is GPT-5.6 Luna (\texttt{gpt-5.6-luna}) with medium
reasoning effort.

The descriptive baseline covers \LiberoTaskCount{} tasks, each evaluated on
\LiberoSeedCount{} seeds, for \LiberoEpisodeDenominator{} episodes. Every cell
uses the same suite-specific budget in read-only evaluation mode, loads no
experience generated during evaluation, and allows neither human nor agent
assistance. Success is accepted only from an official positive LIBERO reward
in a trusted environment receipt. The evaluation manifest freezes the task
catalog, models and prompts, Tool contracts, policy tree, object memory,
budgets, and source-record hashes.
This fixed task--seed protocol requires a completion receipt before aggregate
results are rendered; it contains no task-qualification stage.
Appendix~\ref{app:experiment-protocol} explains the full protocol and how we
mark episodes affected by simulator failures. Appendix~\ref{app:task-results}
lists each task result and the trace used to verify it.
We place LIBERO-Pro outside the formal scope and do not extrapolate robustness
to it; a future evaluation requires its own preregistered manifest.

\subsubsection{OpenETA for Codex configuration}
\label{sec:minimal-geometric-interface}

We evaluate OpenETA for Codex on 130 LIBERO tasks: the four standard suites
(Spatial, Object, Goal, and LIBERO-10, 10 tasks each) and LIBERO-90. The Agent
uses only three embodied Tools: \tool{observe}, \tool{mark_point}, and
\tool{move_to}.

\paragraph{Design motivation.}
Coding agents solve many tasks through a small set of stable operations, such
as file access, editing, shell execution, search, and version control.
OpenETA for Codex applies the same idea to embodied control. It keeps the physical
interface small and leaves task decomposition, target selection, and spatial
reasoning to the multimodal Planner.

The interaction loop needs three Tools. \tool{observe} returns live images.
\tool{mark_point} maps a point selected in an image to a 3D World coordinate.
\tool{move_to} moves the gripper to a target pose. Evaluator lifecycle calls do
not add perception or manipulation capability.

\paragraph{\tool{observe}.}
The Agent requests only the views needed for its next decision. In LIBERO, it
can request a fixed third-person view, a wrist view, or orthographic views along
the World X, Y, and Z axes. Each image has \(512\times512\) resolution.

\paragraph{\tool{mark_point}.}
This Tool turns the Agent's 2D point selection into a 3D coordinate. It supports
two modes.

\textbf{Multi-view mode.}
The Agent first selects a pixel in one orthographic view. The Tool returns the
other two views and draws the corresponding projected ray. The Agent selects a
second point on one of these rays, which resolves the 3D World coordinate.

\textbf{Single-view mode.}
For a point on a visible object surface, the Agent selects one pixel in a
camera view. The Tool returns the first surface intersection along that camera
ray. This mode is sufficient for many pick-and-place tasks.

In both modes, the Tool returns the World XYZ coordinate and an image that marks
the selected point.

\paragraph{\tool{move_to}.}
This Tool moves the gripper to a target position and orientation. The Agent
specifies orientation with two vectors: an approach direction and a jaw
direction. These vectors are easier to interpret than raw rotation angles.

The Tool accepts either an absolute target or a relative change from the
current pose. After each action, it returns the gripper aperture. The Agent can
use this value to check whether the gripper holds an object. Before a close
command, the Tool also renders the target gripper pose. The Agent can inspect
this preview and adjust the pose before execution.

Three evaluator-owned lifecycle calls,
\tool{report_issue}, \tool{check_task} and \tool{finish_episode}, expose no additional perception or
manipulation capability.
We compare GPT-5.6 Luna, GPT-5.6 Terra, and GPT-5.6 Sol at medium reasoning
effort. All three
use the same startup prompt, Tool schemas and result formats, visual feedback
contract, task catalog, and ordered seeds. We report task-level
\textsc{Pass}@\(k\) over seeds 0--4, and native LIBERO task checkers provide the
success verdict.

\subsection{Results and failure modes}

\subsubsection{Full OpenETA baseline}

\ifdefstring{\LiberoResultStatus}{frozen}
  {
The frozen evaluation records 56/400 episode successes (14.0\%) on the complete matrix of 40 tasks and 10 preregistered seeds per task. Every task--seed cell is run once, with no qualification filter or post-hoc rerun.
3 LIBERO-10 episodes lack complete result records because simulator handles expired after approximately 1800 seconds. The batch is therefore a complete descriptive baseline but does not satisfy a strict infrastructure-clean standard. The primary result remains 56/400; 56/397=14.11\% after excluding the 3 cells is diagnostic only and cannot replace the preregistered 400-episode metric.

\begin{table}[htbp]
  \centering
  \small
  \caption{\textbf{OpenETA episode success on the fixed LIBERO matrix.} The Planner is GPT-5.6 Luna (\texttt{gpt-5.6-luna}) with medium reasoning effort.}
  \label{tab:libero-results}
  \begin{tabular}{@{}lccc@{}}
    \toprule
    \textbf{Suite} & \textbf{Success/denom.} & \textbf{Rate} & \textbf{Mean turns (resource \(n\))} \\
    \midrule
    Spatial & 8/100 & 8.0\% & 47.5 (100) \\
    Object & 26/100 & 26.0\% & 31.2 (100) \\
    Goal & 21/100 & 21.0\% & 36.8 (100) \\
    Long / LIBERO-10 & 1/100 & 1.0\% & 58.5 (97) \\
    \midrule
    \textbf{Overall} & \textbf{56/400} & \textbf{14.0\%} & \textbf{43.4 (397)} \\
    \bottomrule
  \end{tabular}
\end{table}

    \ifdefstring{\LiberoProtocolKind}{fixed-matrix}
      {
\paragraph{Stratified descriptive analysis.}
Suite-level episode-success estimates range from 1.0\% (Long / LIBERO-10) to 26.0\% (Object), a descriptive range of 25.0 percentage points. Of 40 tasks, 22 score 0/10, 0 score 10/10, and 18 have mixed outcomes across seeds; 18/40 tasks therefore succeed at least once.
Batch wall time is 19.59 hours. Complete per-episode resource records exist for 397/400 cells and sum to 17,231 turns, 17,141 Tool calls, and approximately 228.2M tokens. Observed peak trace concurrency is 10, peak provider concurrency is 2, and provider queue timeouts are 0.

\begin{table}[htbp]
  \centering
  \small
  \caption{\textbf{Resources stratified by final episode verdict.} Only episodes with complete resource fields are included; wall time is the episode usage elapsed value.}
  \label{tab:libero-resource-strata}
  \begin{tabular}{@{}lrrrr@{}}
    \toprule
    \textbf{Stratum} & \textbf{Complete \(n\)} & \textbf{Mean turns} & \textbf{Mean Tool calls} & \textbf{Mean wall time (s)} \\
    \midrule
    Success & 56 & 30.7 & 30.7 & 1045.0 \\
    Failure & 341 & 45.5 & 45.2 & 1866.6 \\
    \bottomrule
  \end{tabular}
\end{table}

Among 344 failed episodes, the most common mutually exclusive terminal label is \path{episode_timeout} (215/344, 62.5\%). 3 simulator unknown-handle cells are infrastructure contamination and remain zeros in the standard 400-episode denominator; the 56/397=14.11\% exclusion diagnostic does not replace the primary metric. These strata locate computational burden and terminal stage. Early stopping, budget exhaustion, and task difficulty jointly affect resources, so mean-resource differences are not causal effects of computation on success. Suites cover different tasks, and their descriptive range is not a paired significance test.

The categories cover mutually exclusive terminal reasons for all 400 preregistered cells. \texttt{simulator\_unknown\_handle} is marked separately as infrastructure contamination; the others identify where task execution stopped. Tool-return success, stage reachability, and post-hoc observation are not promoted to task success, and a terminal label alone is not treated as an evidenced root cause.

\begin{table}[htbp]
  \centering
  \small
  \caption{\textbf{Mutually exclusive terminal failures in the fixed LIBERO matrix.}}
  \label{tab:libero-failure-taxonomy}
  \begin{tabular}{@{}lrrl@{}}
    \toprule
    \textbf{Terminal class} & \textbf{Episodes} & \textbf{Share of failures} & \textbf{Type} \\
    \midrule
    \texttt{episode\_timeout} & 215 & 62.5\% & Task terminal \\
    \texttt{unattended\_ask\_human} & 66 & 19.2\% & Task terminal \\
    \texttt{max\_turns} & 35 & 10.2\% & Task terminal \\
    \texttt{status\_report\_without\_reward} & 24 & 7.0\% & Task terminal \\
    \texttt{simulator\_unknown\_handle} & 3 & 0.9\% & Infrastructure \\
    \texttt{remote\_episode\_terminated\_without\_reward} & 1 & 0.3\% & Task terminal \\
    \midrule
    \textbf{All failures} & \textbf{344} & \textbf{100.0\%} & -- \\
    \bottomrule
  \end{tabular}
\end{table}

      }
      {}
  }
  {
    \begin{table}[htbp]
      \centering
      \caption{\textbf{Status of formal LIBERO results. Obsolete placeholder
      values are not rendered.}}
      \label{tab:libero-results}
      \begin{tabular}{@{}ll@{}}
        \toprule
        \textbf{Item} & \textbf{Status} \\
        \midrule
        Formal batch & New experiments in progress \\
        Protocol & Pending completion receipt and final manifest \\
        Primary metric & Episode success on the frozen denominator \\
        Per-task evidence & Pending receipt and content-hash freeze \\
        \bottomrule
      \end{tabular}
    \end{table}
  }

Because \LiberoInfrastructureInvalid{} cells are affected by simulator TTL
expiration, the exclusion diagnostic uses
\LiberoEpisodeSuccesses/\LiberoDiagnosticDenominator{} episodes. It cannot replace
the preregistered \LiberoEpisodeDenominator{}-episode descriptive result.

The system remains slow and expensive. Foundation-model inference, perception,
verification, and simulation latency accumulate over long episodes. Among the
\LiberoEpisodeDenominator{} terminal outcomes, \tool{episode_timeout} is the
dominant failure label, followed by unattended \tool{ask_human}, maximum turns,
and status reports without official reward. Every success has an official
positive LIBERO reward, and actual human or agent assistance is zero. The 66
\tool{ask_human} outcomes therefore record unavailable help requests retained
as failures, not assisted results.

These mutually exclusive terminal labels identify where a trajectory stopped;
they do not by themselves establish a physical root cause. Object and Goal
have higher point estimates than Spatial, while Long / LIBERO-10 has one
success. Together with timeout dominance and only
\LiberoTasksWithSuccess/\LiberoTaskCount{} tasks succeeding at least once, the
evidence points to long-horizon subgoal tracking, perceptual rechecking,
placement relations, and remaining-budget coordination as important
bottlenecks. When ambiguity causes a safe stop or help request, OpenETA retains
the last trusted state, unresolved obligations, and stop condition. This
distinguishes safe abstention from unexplained termination, but never
relabels abstention as task success. Appendix~\ref{app:failure-evidence}
defines the failure taxonomy and evidence requirements.

\subsubsection{OpenETA for Codex results}
\label{sec:minimal-breadth-eval}

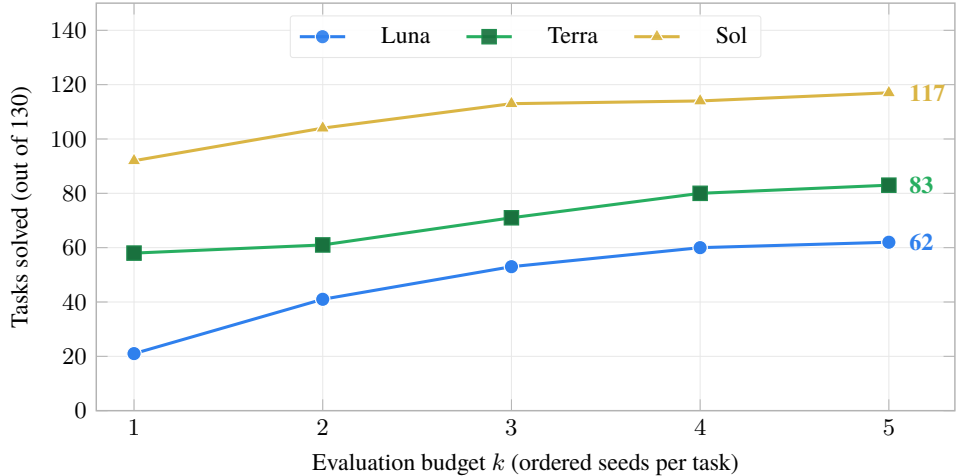
\begin{figure}[htbp]
  \centering
  \begin{tikzpicture}
    \begin{axis}[
      width=0.78\textwidth,
      height=0.42\textwidth,
      xlabel={Evaluation budget $k$ (ordered seeds per task)},
      ylabel={Tasks solved (out of 130)},
      xmin=0.8, xmax=5.35,
      ymin=0, ymax=150,
      xtick={1,2,3,4,5},
      ytick={0,20,40,60,80,100,120,140},
      xmajorgrids=true,
      ymajorgrids=true,
      major grid style={draw=gray!18, line width=0.35pt},
      axis line style={gray!60, line width=0.55pt},
      tick style={gray!60},
      tick label style={font=\small},
      label style={font=\small},
      legend columns=-1,
      legend style={
        draw=gray!20, rounded corners=1pt, fill=white, fill opacity=0.92,
        text opacity=1, at={(0.5,0.97)}, anchor=north,
        font=\small, column sep=1.1em,
      },
      legend cell align={left},
      clip=false,
    ]
    \addplot+[LunaBlue, mark=*, mark size=2.7pt, mark options={fill=LunaBlue, draw=white, line width=0.45pt}, line width=1.15pt]
      coordinates {(1,21) (2,41) (3,53) (4,60) (5,62)};
    \addlegendentry{Luna}
    \node[anchor=west, font=\small\bfseries, text=LunaBlue]
      at (axis cs:5.06,62) {62};
    \addplot+[TerraGreen, mark=square*, mark size=2.7pt, mark options={fill=TerraGreen!65!black, draw=TerraGreen!80!black, line width=0.45pt}, line width=1.15pt]
      coordinates {(1,58) (2,61) (3,71) (4,80) (5,83)};
    \addlegendentry{Terra}
    \node[anchor=west, font=\small\bfseries, text=TerraGreen]
      at (axis cs:5.06,83) {83};
    \addplot+[SolYellow!90!black, mark=triangle*, mark size=3.0pt, mark options={fill=SolYellow!90!black, draw=white, line width=0.45pt}, line width=1.15pt]
      coordinates {(1,92) (2,104) (3,113) (4,114) (5,117)};
    \addlegendentry{Sol}
    \node[anchor=west, font=\small\bfseries, text=SolYellow!90!black]
      at (axis cs:5.06,117) {117};
    \end{axis}
  \end{tikzpicture}
  \caption{\textbf{OpenETA for Codex task-level Pass@\(k\) across 130 LIBERO tasks.}
  GPT-5.6 Luna, GPT-5.6 Terra, and GPT-5.6 Sol use the same Tool and visual
  contracts, medium reasoning effort, and ordered seeds. Pass@\(k\) counts a task as
  solved if any of its first \(k\) ordered-seed episodes succeeds.}
  \label{fig:minimal-passk}
\end{figure}

Figure~\ref{fig:minimal-passk} shows how task coverage accumulates across the
five ordered evaluation seeds.
At \textsc{Pass}@1, Luna, Terra, and Sol solve
\MinimalLunaPassOneSuccesses{}, \MinimalTerraPassOneSuccesses{}, and
\MinimalSolPassOneSuccesses{} tasks, respectively. Coverage rises to
\MinimalLunaPassFiveSuccesses{} (\MinimalLunaPassFivePercent\%),
\MinimalTerraPassFiveSuccesses{} (\MinimalTerraPassFivePercent\%), and
\MinimalSolPassFiveSuccesses{} (\MinimalSolPassFivePercent\%) at
\textsc{Pass}@5. Performance rises with Planner strength under the same Tool
interface. Sol performs especially well on tasks that require precise grasp-point
localization, where it produces fewer empty grasps than Terra and Luna. It also
uses the \tool{move_to} preview more often before acting. Appendix~\ref{app:experiment-protocol}
reports the suite-level \textsc{Pass}@k results.

\subsubsection{Qualitative trace visualization for openeta}

\begin{figure}[htbp]
  \centering
  \includegraphics[width=\textwidth]{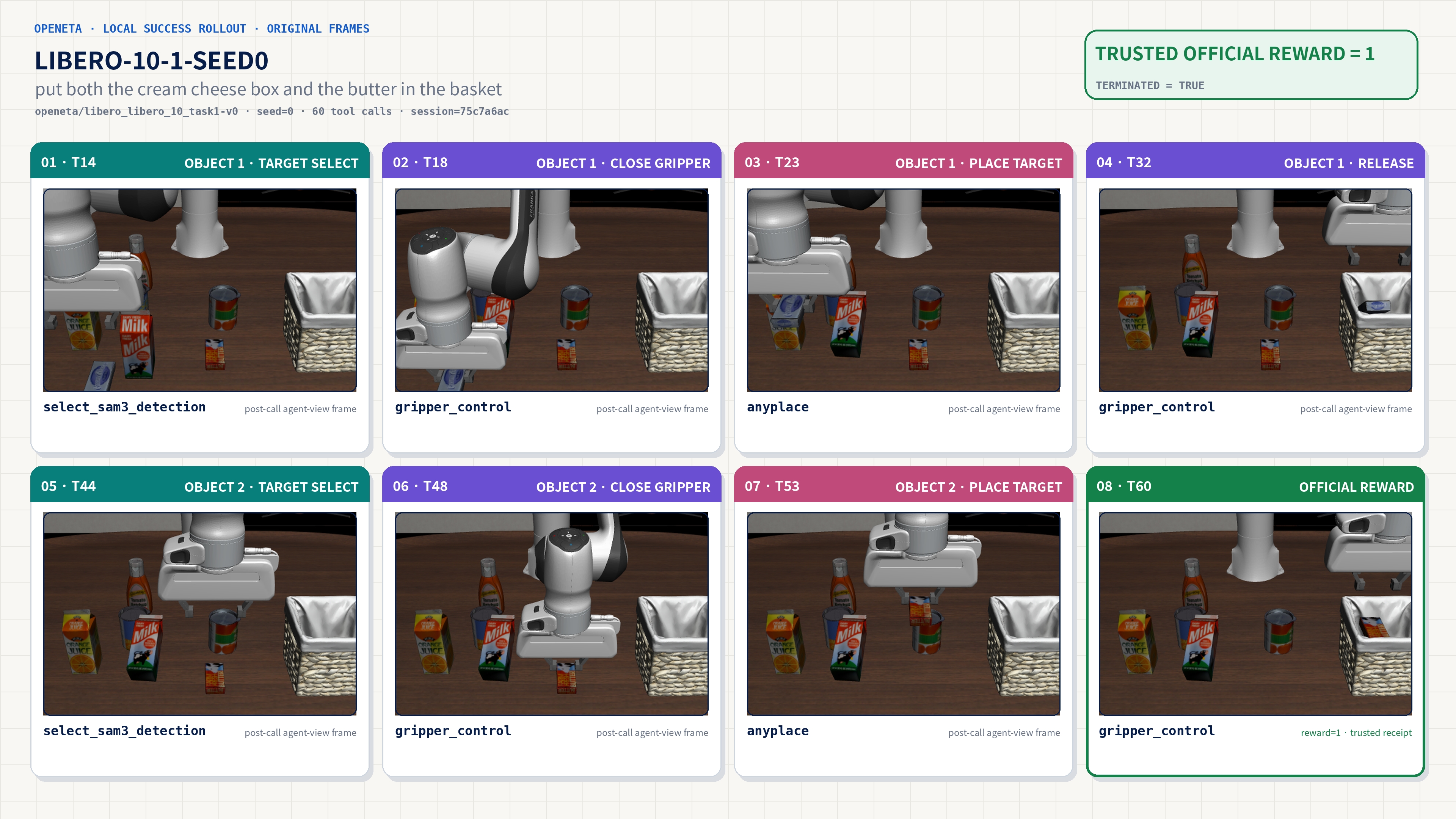}
  \caption{\textbf{Qualitative local LIBERO-10 success trajectory.}
  The two-object rollout exposes target selection, post-action observation,
  placement, release, and trusted reward as separate decision boundaries.
  It visualizes runtime semantics only and is not part of the frozen
  400-episode denominator.}
  \label{fig:qualitative-libero-long}
\end{figure}

Figure~\ref{fig:qualitative-libero-long} shows how the same Tool-mediated
closed loop appears in a successful local LIBERO rollout: each panel is a
post-call observation, the next decision is made from the updated scene, and
completion is accepted only with a trusted official reward. This plate is a
qualitative interface demonstration, not a cell from the frozen fixed matrix;
it contributes neither a success nor a failure to
Table~\ref{tab:libero-results}. Additional local success plates appear in
Appendix~\ref{app:task-results}.

\FloatBarrier

\section{Constrained Agent Self-Evolution}
\label{sec:self_revolution}

OpenETA records observations, Tool calls, task state, environment receipts, and
failure evidence in every trajectory. Agent self-evolution asks whether this
experience can improve later behavior. The model may propose an update, but it
cannot rewrite Tools, controllers, or safety rules. The runtime tests each
update before it can affect future execution. This design builds on reflection,
experience retrieval, and skill libraries
\citep{DBLP:conf/nips/ShinnCGNY23,DBLP:conf/aaai/Zhao0XLLH24,
DBLP:journals/tmlr/WangX0MXZFA24}, but adds checks required for physical action.

\subsection{Notation}

Let \(T\) denote trusted trajectories and \(C\) the fixed Tool and safety
contracts. The current Skill or strategy is \(S\), and a proposed update is
\(u\). A paired evaluation manifest \(M\) fixes the task, seed, budget, and
comparison protocol. The runtime promotes \(u\) only if it passes every gate
defined below.

\subsection{Method}

We test three forms of update, from broad to narrow. A task-local Skill edits
guidance for the full pick-and-place workflow. An exact-task playbook is
extracted from a successful trajectory and loaded only when the environment,
suite, task index, and task text all match. A stage-local delta compares a
successful and a failed trajectory from the same task. It records one symbolic
trigger and one change in action for a specific stage.

No update may store world coordinates, reuse historical poses, redefine atomic
Tools, or bypass safety checks. It also cannot bypass target selection,
attachment verification, placement, release, or official reward. Candidate
generation and candidate evaluation use separate contexts.

\begin{table}[htbp]
  \centering
  \small
  \renewcommand{\arraystretch}{1.12}
  \caption{\textbf{Constrained candidate generation and promotion.}}
  \label{alg:self-evolution-promotion}
  \begin{tabularx}{\textwidth}{@{}rX@{}}
    \toprule
    \multicolumn{2}{@{}l}{\textbf{Input:} trusted trajectories \(T\), fixed
    contracts \(C\), base capability \(S\), paired manifest \(M\)} \\
    \multicolumn{2}{@{}l}{\textbf{Output:} promoted candidate \(u\) or an
    auditable rejection record} \\
    \midrule
    1 & Generate \(u\) from \(T\) and \(S\) without access to evaluation results. \\
    2 & Reject \(u\) if it violates the schema or any invariant in \(C\). \\
    3 & Review \(u\) in a context isolated from candidate generation. \\
    4 & Reject \(u\) if the review does not accept it. \\
    5 & Replay \(S\) and \(u\) on the same task and seed from \(M\). \\
    6 & Reject \(u\) if it cannot reproduce trusted task success. \\
    7 & Compare \(S\) and \(u\) on paired held-out tasks from \(M\). \\
    8 & Reject \(u\) if it shows no objective gain. \\
    9 & Reject \(u\) if it introduces a safety or contract regression. \\
    10 & Promote \(u\) with provenance, version, and rollback metadata. \\
    \bottomrule
  \end{tabularx}
\end{table}

The model can generate a candidate, but only the OpenETA runtime can promote
it. The runtime first checks the candidate, then reproduces the original task,
and finally tests held-out tasks for gains and regressions. This separation
prevents a model-written ``lesson learned'' from gaining execution authority
without evidence.

Formally, the promotion rule is
\begin{equation}
  \mathrm{Promote}(u)
  = D(u) \land Q(u) \land R(u) \land H(u),
\end{equation}
where \(D\) is the deterministic contract check, \(Q\) is the isolated review,
\(R\) is same-task replay, and \(H\) is paired held-out evaluation. A candidate
must pass all four gates.

\subsection{Experimental setting}

We study self-evolution on LIBERO Spatial. In every paired comparison, the
baseline and candidate use the same Tool contracts, base Skill, grasp policy,
calibration, task seed, and resource budget. The only difference is whether
the candidate experience is available. Official environment reward is the
success criterion. Attachment, placement, release, timeout, and invariant
events are diagnostic signals; they never replace official success.

The five studies form an adaptive exploratory sequence. Each study was designed
after inspecting the previous one. They are not five independent trials under
one preregistered protocol, so we report their results separately.

\subsection{Results}

\begin{table}[htbp]
  \centering
  \scriptsize
  \renewcommand{\arraystretch}{1.18}
  \caption{\textbf{Summary of OpenETA self-evolution experiments.}
  ``Baseline/candidate'' denotes paired runs without and with candidate
  experience. Success always means official environment reward.}
  \label{tab:self-revolution-results}
  \begin{tabularx}{\textwidth}{
    @{}p{0.18\textwidth}p{0.23\textwidth}p{0.20\textwidth}X@{}}
    \toprule
    \textbf{Update} & \textbf{Evaluation} &
    \textbf{Official success} & \textbf{Main result} \\
    \midrule
    Online multi-round Skill &
    \SelfOnlineTaskCount{} tasks, one seed, three rounds; two runs &
    \(\SelfOnlineOneRoundOne\!\rightarrow\!\SelfOnlineOneRoundTwo
    \!\rightarrow\!\SelfOnlineOneRoundThree/\SelfOnlineTaskCount\);
    \(\SelfOnlineTwoRoundOne\!\rightarrow\!\SelfOnlineTwoRoundTwo
    \!\rightarrow\!\SelfOnlineTwoRoundThree/\SelfOnlineTaskCount\) &
    A later round can find a success, but the gain does not persist. \\

    Task-local Skill &
    \SelfTaskLocalTaskCount{} tasks \(\times\) \SelfTaskLocalSeeds{} held-out
    seeds; \SelfTaskLocalEpisodesTotal{} episodes &
    baseline \(\SelfTaskLocalOriginalSuccess/\SelfTaskLocalEpisodesPerArm\);
    candidate \(\SelfTaskLocalCandidateSuccess/\SelfTaskLocalEpisodesPerArm\) &
    The candidate reaches attachment and placement stages less often. \\

    Exact-task playbook &
    \SelfPlaybookTaskCount{} tasks \(\times\) \SelfPlaybookSeeds{} held-out
    seeds; \SelfPlaybookEpisodesTotal{} episodes &
    baseline \(\SelfPlaybookBaselineSuccess/\SelfPlaybookEpisodesPerArm\);
    candidate \(\SelfPlaybookCandidateSuccess/\SelfPlaybookEpisodesPerArm\) &
    The candidate uses more turns and times out more often. \\

    Stage-local delta v1 &
    \SelfStrategyOnePairCount{} same-seed replay pairs &
    baseline \(\SelfStrategyOneBaselineSuccess/\SelfStrategyOnePairCount\);
    candidate \(\SelfStrategyOneCandidateSuccess/\SelfStrategyOnePairCount\) &
    Neither arm reproduces success, so held-out evaluation does not run. \\

    Contrastive stage-local delta v2 &
    \SelfStrategyTwoPairCount{} valid same-seed replay pair &
    baseline \(\SelfStrategyTwoBaselineSuccess/\SelfStrategyTwoPairCount\);
    candidate \(\SelfStrategyTwoCandidateSuccess/\SelfStrategyTwoPairCount\) &
    The update changes no key decision and adds
    \SelfStrategyTwoCandidateViolations{} premature gripper opening. \\
    \bottomrule
  \end{tabularx}
\end{table}

No candidate passes all promotion gates. Broad Skill edits sometimes change a
single run, but the effect is not stable. The exact-task playbook also performs
worse than its paired baseline. The two stage-local versions fail to reproduce
the source success, so neither reaches held-out evaluation.

The main result is therefore about control, not performance improvement. The
promotion gates keep unsupported, ineffective, or regressive updates out of the
shared capability library. However, the current update methods show no reproducible improvement in task success.
Most candidates add another visual check or recovery rule; they do not change
the underlying perception or control Tools.
These extra steps can increase Planner turns and timeouts without fixing the
failure.

Future work needs updates with clearer causes and effects. The system should
record when a rule applies, whether the Agent follows it, and how the action
differs from the failed trajectory. A gain should count only when it reproduces
the original success and improves paired held-out tasks without new violations.
Appendix~\ref{app:self-evolution} provides per-task results, stage statistics,
resource use, validity limits, and experiment identifiers.

\section{From Simulation Closed Loops to Real Robots}
\label{sec:sim2real}

OpenETA does not bind task-level intelligence to one simulator or robot.
The Agent consumes normalized observations and proposes structured commands;
backend adapters translate observation, motion, grasp, and release Tools into
simulator or robot requests. When backends preserve functional semantics, task
decomposition, working memory, Skills, recovery, and verification remain
reusable.

What transfers is a task-level closed
loop. Target descriptions, Tool contracts, execution gates,
fresh-observation obligations, checking procedures, and trajectory formats can
transfer; cameras, coordinate frames, motion controllers, gripper interfaces,
sensor calibration, and device-specific safety constraints must be replaced
and revalidated.

A typical migration proceeds as follows:
\begin{enumerate}[leftmargin=*]
  \item validate the observe--act--verify loop in simulation;
  \item integrate the robot backend and calibrate frames, cameras, and end
  effectors;
  \item verify each Tool's functional semantics at low speed and within a
  restricted workspace under emergency-stop supervision;
  \item test post-action observation, timeout recovery, and request
  idempotency; and
  \item run complete tasks in a controlled workspace while retaining the same
  evidence-chain format used in simulation.
\end{enumerate}

\subsection{Qualitative Hardware Demonstration}

\begin{figure}[htbp]
  \centering
  \includegraphics[width=\textwidth]{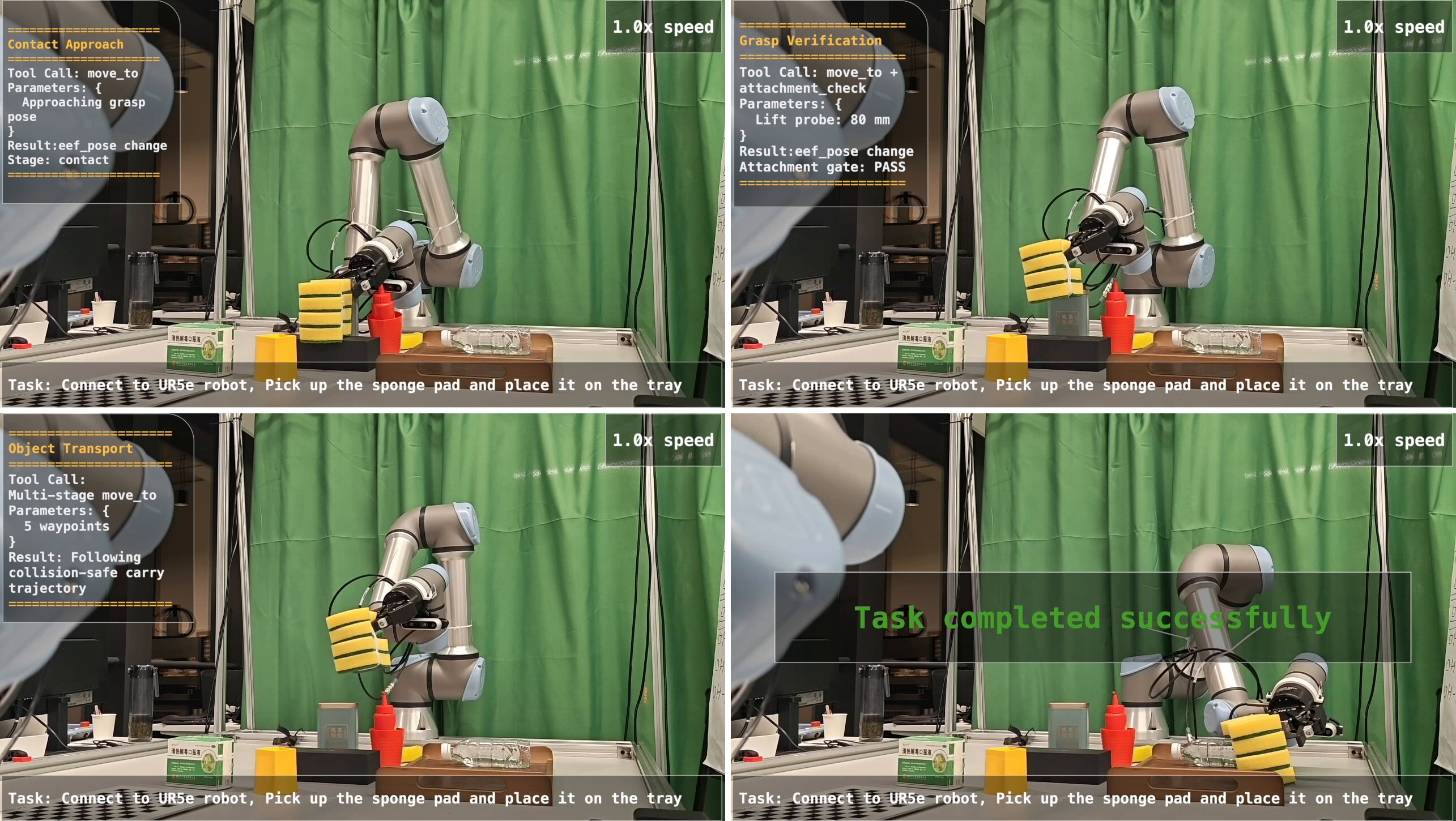}
  \caption{\textbf{Qualitative UR5e sponge-to-tray demonstration.}
  Clockwise from the upper left: contact approach; lift-probe attachment
  verification; final placement on the tray; and multi-waypoint transport.
  The panels are selected frames from one retained successful recording, not
  independent trials or evidence for a success rate.}
  \label{fig:real-robot-sponge-demo}
\end{figure}

Before the formal second-stage study, we recorded two development runs on a
UR5e arm with a Robotiq gripper. One recording
visibly covers a complete sponge-to-tray sequence: contact approach, a lift
probe and attachment check, multi-waypoint transport, and final placement
(Figure~\ref{fig:real-robot-sponge-demo}). The other visibly establishes a
bell-pepper grasp and transport, but its final in-basket relation is not
supported by the archived view. These recordings show that the closed-loop Tool pipeline can
drive the hardware through substantive manipulation stages.

Development testing exposed two coupled Sim2Real bottlenecks. First, depth from the
RealSense D435i wrist and supplementary third-person cameras and the RealSense
L515 main third-person camera was substantially less complete than simulator
depth. Depth-estimation enhancement improved the input but did not reliably
restore geometry sufficient for high-quality pose estimates. Second, some
lateral-grasp target poses were followed by acceleration-limit protective
stops. Trace review suggests controller PD tuning or trajectory shaping as a
possible contributor, but the retained recordings do not establish a causal diagnosis.
Both issues motivate the frozen calibration, depth-validity, motion-limit, and
per-trial evidence checks planned for the second-stage evaluation.

The frozen release evidence remains \textbf{interface integration}. Its source
snapshot does not establish SDK availability, device connectivity, calibration
quality, primitive success, task success, or safety certification; the
development recordings remain qualitative demonstrations.

``Transfer'' does not mean engineering-free adaptation, nor do simulation
results imply physical safety. Collision checking, control frequency,
emergency stop, payload limits, calibration error, and shared-workspace risk
require device-specific validation. Appendix~\ref{app:sim2real-safety}
separates interface integration, primitive validation, and task validation,
and lists the calibration, limits, emergency-stop, idempotency, and supervision
evidence required for each physical batch.

Real-robot demonstrations and updates will appear at:
\begin{center}
  \url{https://github.com/OpenMOSS/OpenETA}
\end{center}

\section{Limitations and Future Work}
\label{sec:future}

OpenETA connects multimodal planning, atomic Tools, textual Skills,
session-level memory, simulation and perception interfaces, parallel
evaluation, and immutable trajectory records in one reproducible loop. The
frozen LIBERO matrix records
\LiberoEpisodeSuccesses/\LiberoEpisodeDenominator{} successful episodes, and
\LiberoTasksWithSuccess/\LiberoTaskCount{} tasks succeed at least once.
Long / LIBERO-10 is the weakest suite. Simulator TTL expiration affects
\LiberoInfrastructureInvalid{} cells. Current limitations include:
\begin{itemize}[leftmargin=*]
  \item timeouts dominate the formal failure labels, and subgoal progress,
  budget, and remaining-time management remain inadequate in multi-object
  tasks;
  \item placement relations, release timing, and attachment stability remain
  major bottlenecks;
  \item bimanual coordination, dynamic contact, and mobile manipulation lack
  mature Tools and checkers;
  \item simulation cannot establish real-robot safety, control frequency,
  emergency-stop behavior, calibration, or payload limits; and
  \item current general-purpose models do not reliably adopt a habit of
  observing and correcting after every physical action.
\end{itemize}

\begin{figure}[htbp]
  \centering
  \includegraphics[width=\textwidth]{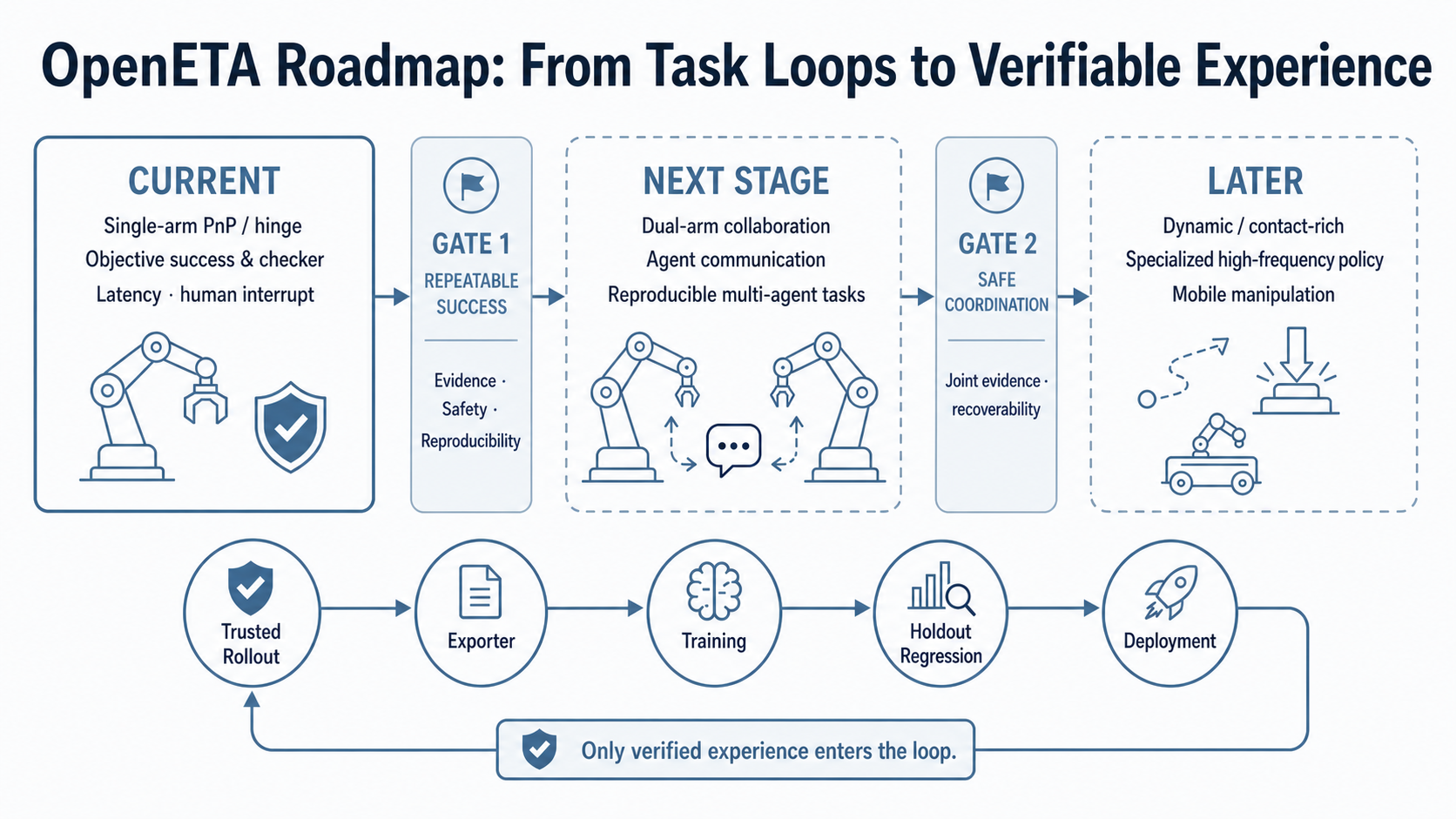}
  \caption{\textbf{OpenETA research roadmap.}
  Progression from current single-arm tasks to multi-agent coordination and
  later contact-rich or mobile manipulation is gated by repeatable success,
  safety, reproducibility, and recoverability. The later stages and training
  loop are research targets, not claims about the current release.}
  \label{fig:openeta-roadmap}
\end{figure}

Next, we will remove simulator-session TTL and resource-record gaps, then rerun
the frozen matrix under an infrastructure-clean protocol. We will reduce
context and Tool-call cost while improving long-horizon memory and recovery.
We will also expand the Tool set toward bimanual, contact-rich, and mobile
tasks. Finally, we will connect trusted rollouts, training, regression tests,
and redeployment in a self-improvement loop.
Figure~\ref{fig:openeta-roadmap} summarizes this staged agenda and the evidence
gates required before progressing to broader embodiments and harder tasks.

We also plan to let the Agent move a free camera around a selected region. This
extra viewpoint can reduce occlusion before grasping or placement. The resulting
interaction traces may also provide training data for spatial reasoning models.

Future VLAs and WAMs can be invoked as high-level ETA capabilities. Specialized
VLAs such as CoRE-VLA and Gemini Robotics
\citep{zhang2026corevlascalablerobustvisionlanguageaction,
geminirobotics2025} may provide short-horizon control for dexterous,
contact-rich, or multi-embodiment tasks that are difficult to express through
general code composition. A WAM may predict future states and compare plan
consequences. Omnimodal embodied-agent
systems further suggest treating smart-home gateways, Web search, interaction,
and memory as capabilities in the same task loop
\citep{DBLP:journals/corr/abs-2606-27251}. The task-level Agent chooses
when to invoke each capability from current evidence, while the Interface
retains execution gating and post-action verification for state-changing
physical calls.

\FloatBarrier

\section{Conclusion}
\label{sec:conclusion}

Embodied intelligence needs stronger models, but scale alone does not create a
trustworthy physical closed loop. A system must continually answer: what was
observed, what may change, who authorizes the change, what actually happened,
and under what conditions the resulting experience remains credible.

ETA realizes these questions as a protocol among Agent, Interface, and World.
OpenETA implements it with atomic Tools, side-effect classes, execution gates,
trusted environment receipts, fresh-observation obligations, and replayable
trajectories.
\ifdefstring{\LiberoResultStatus}{frozen}
  {Without task-specific policy training, the frozen LIBERO matrix achieves
  \LiberoEpisodeSuccesses/\LiberoEpisodeDenominator{} episode successes, and
  \LiberoTasksWithSuccess/\LiberoTaskCount{} tasks succeed at least once.
  Because \LiberoInfrastructureInvalid{} cells are contaminated by simulator
  TTL expiration, this is a complete descriptive baseline rather than a
  strictly infrastructure-clean confirmatory result.}
  {Development trajectories contain long-horizon pick-and-place episodes
  confirmed by official reward, but they are not formal benchmark evidence.}
OpenETA for Codex evaluates the same three-Tool interface with three Planners on
130 LIBERO tasks. At \textsc{Pass}@5, GPT-5.6 Luna, Terra, and Sol solve
\MinimalLunaPassFiveSuccesses{}, \MinimalTerraPassFiveSuccesses{}, and
\MinimalSolPassFiveSuccesses{} tasks, respectively. Sol also solves
\MinimalSolPassOneSuccesses{} tasks on the first seed. These results show that
stronger general-purpose Planners can use the same small physical interface
more effectively.
\ifdefstring{\RealRobotResultStatus}{frozen}
  {The audited real-robot batch additionally records
  \RealRobotAutonomousSuccesses/\RealRobotValidTrials{} fully autonomous
  successes (\RealRobotAutonomousRatePercent\%) at the stated
  physical evidence level.}
  {The real-robot backend currently has interface-level integration only; it
  supports no claim of primitive execution, task execution, or real-robot
  audited success rate. Retained qualitative recordings show a complete
  sponge-to-tray example and expose engineering priorities, but do not raise
  that formal evidence level.}

No self-evolution candidate passed the promotion gate. The evidence therefore
supports the mechanism claim that experience should affect execution only
after paired non-regression validation, not a performance-improvement claim.
Simulation failures and interface-integration work jointly expose current
limits in localization, placement, latency, and physical safety.

When every observation, command, action, and result leaves a structured record,
an embodied system becomes easier to evaluate, debug, and improve. It can then
learn from verified physical experience as well as from larger foundation
models.

\clearpage
\renewcommand{\bibfont}{\footnotesize}
\bibliographystyle{unsrtnat}
\bibliography{main}

\clearpage
\appendix
\section{A Pick-and-Place Task Is an Evidence Chain}
\label{app:evidence-chain}

Consider ``put the alphabet-soup can in the basket.'' Before grasping, the
robot must determine which scene instance the instruction denotes. In a scene
with similar packages, reducing the query to ``soup can'' can produce a
high-quality mask for the wrong instance.

\subsection{From asset reference to explicit selection}

OpenETA first retrieves a target reference image from controlled object memory,
then aligns reference appearance with the current scene to prompt visual
localization. Retrieval separates object identity from relational context: in
``pick up the black bowl on the cookie box,'' the retrieval term is ``black
bowl,'' while ``on the cookie box'' remains a scene constraint.

Candidate masks rank hypotheses; they do not mean that the target is
confirmed. Every nonempty segmentation result creates an explicit-selection
obligation. The original image, candidate overlay, and stable candidate IDs go
to the visual planner. Grasp estimation and physical control remain blocked
until that obligation is discharged. If no candidate matches, the system
records a structured rejection and returns to localization.

\subsection{From grasp receipt to task truth}

The grasp estimator produces a provenance- and score-bearing candidate queue.
Candidate IDs survive camera-to-world transformation. A safety rejection or
candidate-specific motion failure activates the next candidate, whereas bad
input, calibration errors, and transport failures are not mislabeled as
geometric failures.

Closing the gripper does not prove a grasp. The system still checks whether the
target moves with the end effector, whether its original location is vacated,
and whether the fresh observation matches the selected target, camera
parameters, and scene version. After release, visual plausibility is likewise
insufficient: completion must be grounded in trusted reward, an environment
termination condition, or a task checker.

\begin{figure}[htbp]
  \centering
  \includegraphics[width=\textwidth]{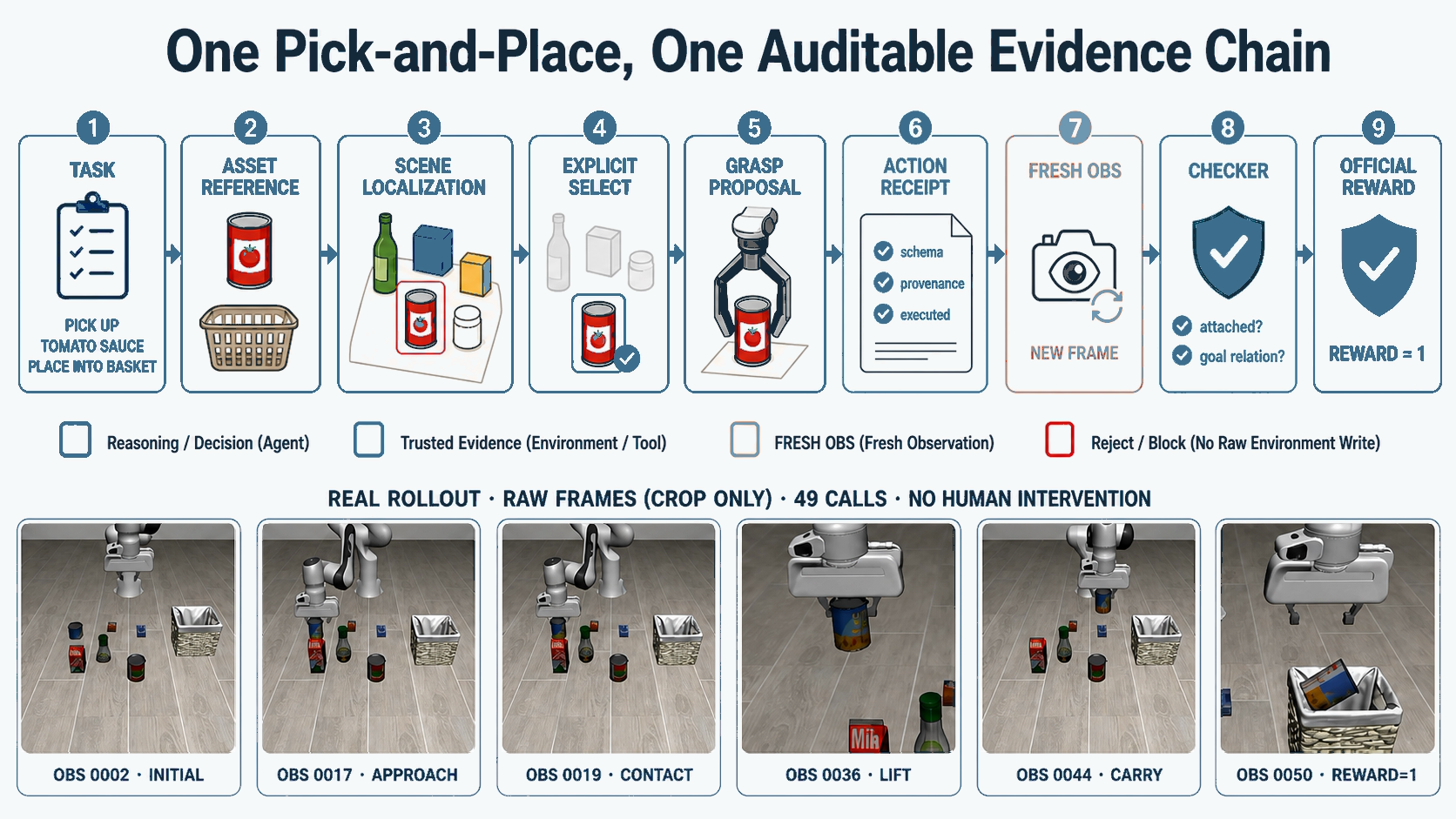}
  \caption{\textbf{An auditable pick-and-place evidence chain.}
  Reasoning decisions, trusted environment evidence, and fresh post-action
  observations are stored separately; the frames show a recorded trajectory
  from initial observation to official reward.}
  \label{fig:evidence-chain}
\end{figure}

An ETA therefore executes the evidence-gated procedure in
Table~\ref{alg:evidence-gated-pick-place}, rather than recording only the
terminal assertion ``pick-and-place succeeded.'' Every world-changing call is
followed by a fresh observation, and only a trusted checker or official
environment reward can terminate the procedure as a success.
Figure~\ref{fig:evidence-chain} visualizes how these reasoning decisions,
post-action observations, and trusted verdicts form one auditable trajectory.

\begin{table}[htbp]
  \centering
  \small
  \renewcommand{\arraystretch}{1.12}
  \caption{\textbf{Evidence-gated pick-and-place execution.}}
  \label{alg:evidence-gated-pick-place}
  \begin{tabularx}{\textwidth}{@{}rX@{}}
    \toprule
    \multicolumn{2}{@{}l}{\textbf{Input:} task (g), observation (o), Tool contracts (C)} \\
    \multicolumn{2}{@{}l}{\textbf{Output:} trusted-success receipt or typed-failure record} \\
    \midrule
    1 & Set \(r\leftarrow\tool{retrieve_asset_reference}(g)\) and
        \(D\leftarrow\tool{localize}(r,o)\). \\
    2 & \textbf{while} \(D\neq\varnothing\) \textbf{do} select
        \(d\leftarrow\mathrm{PlannerSelect}(D,o)\). \\
    3 & \quad \textbf{if} explicit selection of \(d\) is unconfirmed, reject \(d\) and continue. \\
    4 & \quad Generate provenance-bearing grasp candidates
        \(G\leftarrow\mathrm{EstimateGrasps}(d,o)\). \\
    5 & \quad \textbf{for each} \(q\in G\) \textbf{do} execute \(q\) through \(C\), then obtain fresh \(o'\). \\
    6 & \qquad \textbf{if} receipt and \(o'\) disagree, return \tool{receipt_or_scene_mismatch}. \\
    7 & \qquad \textbf{if} attachment is not verified, continue with the next \(q\). \\
    8 & \qquad Place and release through \(C\); obtain fresh \(o''\) and trusted verdict \(v\). \\
    9 & \qquad \textbf{if} \(v.\mathrm{official\_reward}>0\), return trusted success. \\
    10 & \qquad \textbf{else} return \tool{goal_not_verified}. \\
    11 & \textbf{return} \tool{no_verified_grasp_candidate}. \\
    \bottomrule
  \end{tabularx}
\end{table}

Such trajectories support not only debugging but also structured data export,
training, and regression evaluation. Successes, failures, candidate switches,
recoveries, and checker decisions become filterable and reusable experience.


\FloatBarrier

\section{Reproducible Experimental Protocol}
\label{app:experiment-protocol}

This appendix defines the minimum reporting contract used by the experiments.
It does not prescribe one machine or budget; it specifies what must be frozen
and disclosed so that comparisons across models, experience versions, and
backends remain auditable.

\subsection{Evaluation unit and success definition}

An \emph{episode} is the complete interaction in one initialized environment,
from the first trusted observation to official success, environment
termination, resource exhaustion, safety stop, or infrastructure failure. It
must bind task ID and text, environment version, seed, treatment arm, model and
prompt, Skill/strategy version, Tool-registry version, and resource budget.

Success requires a host-accepted trusted environment reward or equivalent
official task verdict. Tool-call success, reaching a pose, attachment PASS, or
an Agent completion report are stage evidence only. For task \(i\),
\begin{equation}
  S_i^{@k}=\max_{1\leq j\leq k} r_{i,j},\qquad
  \mathrm{Pass@}k=\frac{1}{N}\sum_{i=1}^{N}S_i^{@k},
\end{equation}
where \(r_{i,j}\in\{0,1\}\) is the official verdict for seed \(j\). A
Pass@\(k\) report must also state the total episode count.

We distinguish two episode-level designs. A fixed matrix evaluates the
Cartesian product of all \(N\) tasks and a seed set \(\mathcal{S}\):
\begin{equation}
  \mathrm{Success}_{\mathrm{fixed}}
  =\frac{\sum_{i=1}^{N}\sum_{s\in\mathcal{S}}r_{i,s}}
  {N|\mathcal{S}|}.
\end{equation}
A two-stage design first uses seed 0 to qualify a trusted frozen playbook, then
runs preregistered held-out seeds only for qualified tasks; unqualified tasks
remain zero in the full formal denominator. ``Held-out'' is valid only when
the manifest proves those seeds did not contribute to strategy formation.
Fixed-matrix success, two-stage episode success, and task-level Pass@\(k\)
answer different questions and are not interchangeable.

\subsection{OpenETA for Codex setup}

Since Libero's fixed single third-person perspective is not convenient for synthesizing complete point clouds, we have added multiple camera views in Libero to synthesize high-quality point clouds. These are then used in \tool{Observe} as three-view feedback to the agent, facilitating agent interaction.

The OpenETA for Codex comparison uses 512-by-512 images, live multi-view point
selection, a 5,000-step simulator horizon, and a 5,400-second attempt timeout.
For every planner, tasks receive ordered seeds 0--4 and stop after the first
valid native-checker success. The task catalog, startup prompt, Tool and visual
contracts, image configuration, and budgets are fixed across Luna, Terra, and
Sol; each planner uses medium reasoning effort. Our main results—specifically, the success rates on 4‑suite + libero‑90—are shown in Table~\ref{tab:minimal-passk-suite}.

\begin{table}[htbp]
  \centering
  \scriptsize
  \renewcommand{\arraystretch}{1.08}
  \caption{\textbf{OpenETA for Codex successes by Planner, suite, and Pass@\(k\).}
  GPT-5.6 Luna, Terra, and Sol use medium reasoning effort. Each cell counts
  tasks solved within the first \(k\) ordered seeds.}
  \label{tab:minimal-passk-suite}
  \begin{tabular}{@{}llrrrrrr@{}}
    \toprule
    \textbf{Planner} & \textbf{Suite} & \textbf{Tasks} & \textbf{P@1} & \textbf{P@2} &
    \textbf{P@3} & \textbf{P@4} & \textbf{P@5} \\
    \midrule
    Luna & Spatial & 10 & 4 & 7 & 8 & 9 & 9 \\
     & Object & 10 & 1 & 2 & 2 & 3 & 3 \\
     & Goal & 10 & 1 & 3 & 5 & 5 & 6 \\
     & Long / LIBERO-10 & 10 & 0 & 1 & 2 & 2 & 2 \\
     & LIBERO-90 & 90 & 15 & 28 & 36 & 41 & 42 \\
    & Total & 130 & 21 & 41 & 53 & 60 & 62 \\
    \midrule
    Terra & Spatial & 10 & 8 & 8 & 8 & 10 & 10 \\
     & Object & 10 & 1 & 2 & 4 & 4 & 6 \\
     & Goal & 10 & 6 & 6 & 6 & 6 & 6 \\
     & Long / LIBERO-10 & 10 & 2 & 2 & 3 & 3 & 3 \\
     & LIBERO-90 & 90 & 41 & 43 & 50 & 57 & 58 \\
    & Total & 130 & 58 & 61 & 71 & 80 & 83 \\
    \midrule
    Sol & Spatial & 10 & 10 & 10 & 10 & 10 & 10 \\
     & Object & 10 & 8 & 9 & 10 & 10 & 10 \\
     & Goal & 10 & 8 & 8 & 8 & 8 & 8 \\
     & Long / LIBERO-10 & 10 & 7 & 8 & 10 & 10 & 10 \\
     & LIBERO-90 & 90 & 59 & 69 & 75 & 76 & 79 \\
    & Total & 130 & 92 & 104 & 113 & 114 & 117 \\
    \bottomrule
  \end{tabular}
\end{table}

\subsection{Frozen manifest fields}

\begin{table}[htbp]
  \centering
  \small
  \renewcommand{\arraystretch}{1.15}
  \caption{\textbf{Minimum fields frozen for a formal experimental batch.}}
  \label{tab:experiment-manifest}
  \begin{tabularx}{\textwidth}{@{}p{0.22\textwidth}X@{}}
    \toprule
    \textbf{Category} & \textbf{Required fields} \\
    \midrule
    Code and runtime &
    Release commit, dirty-diff hash, Python/dependency versions, operating
    system, compute device, and simulator/robot backend version. \\
    Model and planner &
    Provider, model ID, inference parameters, system-prompt and planner hashes,
    plus request window if the hosted model is mutable. \\
    Capability configuration &
    Tool registry and visible contracts, Skill/strategy hash, perception,
    grasp, and placement model versions, and calibration/scene versions. \\
    Task and randomness &
    Benchmark version, suite, task ID and normalized text, seed, reconstructible
    initial-state reference, and attempt index. \\
    Resource budget &
    Maximum planner turns, Tool calls, wall time, token or monetary budget, and
    timeout thresholds; paired arms use equal budgets. \\
    Treatment &
    The sole allowed difference between arms, such as an exact-task playbook
    hash; manifest comparison verifies every other frozen field. \\
    Outputs and evidence &
    Official reward, terminal reason, stages, violations, failure class,
    resources, and content hash or stable index for the rollout bundle. \\
    \bottomrule
  \end{tabularx}
\end{table}

The self-evolution studies predate this release-level contract. Their source
manifests establish task, seed, common budget, official reward, and treatment,
but do not uniformly freeze the primary planner ID, system-prompt hash, or
Skill-bundle hash. This does not alter recorded rewards or paired counts, but
limits model attribution and third-party reproduction.

\subsection{Batch validity and exclusion}

An episode is \emph{infrastructure-invalid} if environment creation fails,
task or seed identity mismatches, a required service is unavailable, treatment
is not loaded, records/receipts are missing, or a frozen field drifts. Such an
episode is not evidence of task incapability. Exclusion rules and the primary
denominator must be fixed before seeing results. If the preregistered primary
metric includes all launched matrix cells, later contamination remains in the
raw denominator and a contamination-excluded value is reported only as a
diagnostic. If one arm of a pair may be affected, the whole pair is invalid.

Exhausting time, calls, tokens, or allowed recoveries in a valid environment is
a system failure under the stated budget, not infrastructure invalidity.
Reports separately count valid and invalid episodes, successes, resource
exhaustion, safety stops, and other task failures.

\subsection{Paired comparison and uncertainty}

Experience, Skill, and strategy comparisons use the same task, seed, initial
condition, model, Tools, prompt, budget, and backend; only treatment changes.
Binary rates report their numerator and denominator. Paired outcomes include
baseline-only, candidate-only, both-success, and both-fail, with an exact
McNemar test when sample size permits. Stage reachability, turns, Tool calls,
tokens, time, and invariant violations remain diagnostics and never replace
the preregistered primary metric.

\subsection{Replayable evidence bundle}

Each release-grade episode has a manifest plus append-only model-call,
Tool-call, transition, episode-summary, and artifact-index records. Large
images, point clouds, and videos use content-addressed references. Before
release, credentials, tokens, user paths, device identifiers, and unnecessary
provider payloads are removed while preserving fields needed to audit reward
and failure classification.

\subsection{Automated audit and import gates}

The two-stage importer \path{scripts/import_libero_task_eval.py} accepts only a
completed \path{openeta.libero_task_eval.v1} batch with
\path{run_manifest.json}, \path{task_catalog.json}, \path{summary.json}, and
\path{COMPLETED.json}. It verifies receipt hashes, four fixed suites with ten
tasks each, task text and environment IDs, held-out seeds, arithmetic, cleanup,
and infrastructure health. It also freezes planner/provider identity,
timeouts, retry limits, Tool and task-catalog fingerprints, SAM3 and grasp
backend contracts, object-reference bundles, and the MolmoPoint model revision
when configured. Service URLs, local paths, sessions, and handles are omitted
from the public snapshot.

The fixed-matrix importer
\path{scripts/import_libero_fixed_matrix_eval.py} separately audits the full
four-suite, ten-task, seed-0--9 Cartesian product. It validates per-session
records, batch outcomes, and machine audit summary; checks task identity,
read-only/no-evolution mode, budgets, planner and prompt hashes, and resources;
and accepts success only from an official binary reward in a trusted receipt.
An initialized trajectory that safely stops before any reward-bearing action
remains a zero in the fixed denominator. Unattended help requests also remain
zero; any actual human guidance or agent assistance rejects the batch.

Audit occurs in two stages. First, every record is traversed to construct an
independent completion receipt containing manifest, preflight, provenance,
session index, outcome, audit summary, and episode hashes. Second, hashes are
recomputed before producing the sanitized snapshot, main
table and plot, success/failure resource diagnostics, mutually exclusive
terminal distribution, deterministic representative failures, per-task
tables, and reproducibility settings.

The frozen batch contains 397 complete session results and three cells with
initialization evidence only. These are accepted only because both batch
outcomes and the machine audit identify simulator unknown-handle expiration
with matching task, seed, budget, and no assistance. They count as zero in the
preregistered 400-episode primary denominator and as infrastructure
contamination; resources cover only
\LiberoResourceEpisodeCount{} complete records. The
\LiberoEpisodeSuccesses/\LiberoDiagnosticDenominator{} contamination-excluded
value is diagnostic only and cannot replace the primary metric. The
\tool{freeze_libero_fixed_report.py} staging workflow regenerates the snapshot
and eight \LaTeX{} products, updates manifests and rollout indices, rebuilds
the content index, and runs the publication audit before writing back.

\ifdefstring{\LiberoResultStatus}{frozen}
  {
    \ifdefstring{\LiberoProtocolKind}{fixed-matrix}
      {
\subsection{Reproducibility Settings for the Frozen LIBERO Batch}
\label{app:libero-reproducibility}

Displayed hashes are 12-character prefixes. Full SHA-256 values, source-file hashes, per-task identity, and per-episode resources remain in the same sanitized audit snapshot. The table excludes service addresses, local paths, session handles, and credentials.

\begin{table}[htbp]
  \centering
  \scriptsize
  \renewcommand{\arraystretch}{1.16}
  \caption{\textbf{Frozen reproducibility settings for the formal LIBERO matrix.}}
  \label{tab:libero-reproducibility}
  \begin{tabularx}{\textwidth}{@{}p{0.24\textwidth}X@{}}
    \toprule
    \textbf{Item} & \textbf{Frozen value} \\
    \midrule
Batch and code & \texttt{libero-fixed-40x10-20260730-r1}; commit \texttt{4481a9ffd3b2\ldots}. \\
Evaluation design & 40 tasks $\times$ 10 seeds = 400 episodes, full Cartesian product; read-only, no human intervention, no online self-evolution. \\
Integrity boundary & 3 infrastructure-contaminated cells; descriptively complete but not strictly infrastructure-clean. \\
Planner & \texttt{AI Gateway} / \texttt{gpt-5.6-luna}; \texttt{medium} reasoning effort; prompt \texttt{e3cea22d277e\ldots}. \\
Simulation contract & Tool contract \texttt{30e90334351e\ldots}; task catalog \texttt{ee93dc8d3bb2\ldots}. \\
Perception contracts & \texttt{anygrasp}=\texttt{ff0fd3f032f3\ldots}; \texttt{anyplace}=\texttt{b9afd7f7ce8e\ldots}; \texttt{contact\_graspnet}=\texttt{0fa6e5bf80d8\ldots}; \texttt{depth\_prior}=\texttt{26e9a311718d\ldots}; \texttt{graspgenx}=\texttt{55ce2fb4f63c\ldots}; \texttt{molmopoint}=\texttt{6e67aec902b5\ldots}; \texttt{sam3}=\texttt{61b459c61c35\ldots}; MolmoPoint capability probe \texttt{true}. \\
Object memory & 3 frozen references; bundle \texttt{0dd76cf8b8d1\ldots}. \\
Strategy and experience & \texttt{fixed\_read\_only\_no\_self\_evolution}; Skill tree \texttt{a5bc791b12f6\ldots}; grasp strategy \texttt{84607b1318a7\ldots}; task playbook \texttt{b89bba287747\ldots}; calibration \texttt{a46ed83b67ba\ldots}. \\
Concurrency and cleanup & trace peak 10/10; provider peak 2/2; queue timeouts 0; active environments after cleanup 0. \\
    \bottomrule
  \end{tabularx}
\end{table}

\begin{table}[htbp]
  \centering
  \scriptsize
  \caption{\textbf{Preregistered episode budgets by LIBERO suite.}}
  \label{tab:libero-suite-budgets}
  \begin{tabular}{@{}lrrrr@{}}
    \toprule
    \textbf{Suite} & \textbf{Max turns} & \textbf{Max calls} & \textbf{Timeout (s)} & \textbf{Max tokens} \\
    \midrule
    Spatial & 100 & 400 & 1800 & 10000000 \\
    Object & 100 & 400 & 1800 & 10000000 \\
    Goal & 100 & 400 & 1800 & 10000000 \\
    Long / LIBERO-10 & 200 & 400 & 3600 & 10000000 \\
    \bottomrule
  \end{tabular}
\end{table}

These are stopping, not exclusion, rules: a valid episode that exhausts any budget remains a task failure.
}
      {}
  }
  {}

The real-robot importer \tool{scripts/import_real_robot_eval.py} applies a
parallel gate to manifests, trials, summaries, and completion receipts. It
freezes clean code and interface-contract hashes, sanitized hardware
configuration, calibration error and valid range, emergency-stop and workspace
evidence, supervision, preregistered success/exclusion rules, planner identity,
and budgets. Valid trials reference continuous video, structured rollout, and
verdict evidence. Physical intervention and safety stops remain autonomous
failures. A claim can rise from interface integration to primitive or task
validation only when both the code commit and interface contract match the
release snapshot.

\subsection{Claim--evidence matrix and public artifact index}

\begin{table}[htbp]
  \centering
  \scriptsize
  \renewcommand{\arraystretch}{1.16}
  \caption{\textbf{Primary claims, minimum evidence, and interpretation limits.}}
  \label{tab:claim-evidence-matrix}
  \begin{tabularx}{\textwidth}{
    @{}p{0.20\textwidth}p{0.34\textwidth}X@{}}
    \toprule
    \textbf{Claim} & \textbf{Minimum supporting evidence} &
    \textbf{What it does not establish} \\
    \midrule
    Release implementation and 44 Tools &
    System-contract snapshot, source, and dependency hashes from a clean
    release commit. &
    Installed dependencies on the evaluation host or any task success. \\
    Formal LIBERO capability &
    Completion receipt, frozen catalog, complete protocol denominator,
    sanitized snapshot, and deterministically generated tables. &
    Exploration, qualification success, and development traces cannot fill the
    formal denominator. \\
    Self-evolution mechanism &
    Frozen snapshots, task and manifest hashes, paired outcomes, stage
    evidence, and promotion verdicts from six audited batches. &
    Zero promotions prove neither mechanism effectiveness nor universal
    failure of textual experience. \\
    Real-robot interface integration &
    Interface, driver registration, gates, and configuration hashes from a
    clean release commit. &
    SDK/device availability, calibration, primitive success, or task success. \\
    Real-robot primitive or task capability &
    Completed preregistered batch, continuous video and rollout hashes,
    verdict evidence, interventions, stops, and invalid trials. &
    Simulation, source presence, or edited success clips cannot raise the
    physical claim level. \\
    Public evidence bundle &
    Publication manifest, content-addressed artifact index, sanitized results,
    generated tables, and audit scripts. &
    A path alone does not prove content stability; a hash does not imply raw
    traces are public. \\
    \bottomrule
  \end{tabularx}
\end{table}

\path{publication_manifest.json} records claim status and evidence paths.
\path{results/report_artifact_index.json} deterministically records each
referenced file's SHA-256, size, and claim membership. Any content or path
change invalidates the old index. The separate
\path{results/sanitized_rollout_index.json} maps stable run/task/trial IDs to
public or restricted trace, video, receipt, or verdict hashes. Restricted
records state why; public records require HTTPS locations. Raw self-evolution
rollouts remain restricted, formal LIBERO has forty content-addressed task
bundles, and formal real-robot results remain pending.

The final archive is built from an allowlist by
\tool{scripts/build_publication_bundle.py}, not by recursively copying the
repository. It includes the PDF, \LaTeX{} source and figures, build/audit
tools, manifests, indices, and named sanitized results. Its
\tool{BUNDLE_MANIFEST.json} fixes file hashes, sizes, archive paths,
timestamps, permissions, order, and compression. Draft bundles retain explicit
provisional warnings; a release bundle requires the PDF, log, content index,
and all release gates to pass.

\section{Per-Task Results and Reporting Format}
\label{app:task-results}

An aggregate rate cannot show which tasks are stable, which depend on a
particular seed, or whether failures occur during perception, grasp,
attachment, or placement. Every formal benchmark batch should therefore
publish machine-readable per-task results and generate paper tables from the
same frozen source.

\begin{table}[htbp]
  \centering
  \small
  \renewcommand{\arraystretch}{1.15}
  \caption{\textbf{Minimum columns for per-task results.}}
  \label{tab:per-task-result-fields}
  \begin{tabularx}{\textwidth}{@{}p{0.25\textwidth}X@{}}
    \toprule
    \textbf{Field} & \textbf{Meaning} \\
    \midrule
    suite / task ID / task text &
    Stable benchmark identity with complete normalized task text. \\
    seeds and attempts &
    Executed seeds, attempt order, and preregistration membership. \\
    metric / denominator &
    Task-level Pass@\(k\), fixed-matrix episode success, or two-stage held-out
    episode success, with binary verdicts and task/episode denominators. \\
    terminal reason &
    Official success, environment termination, resource exhaustion, safety
    stop, task failure, or infrastructure invalidity. \\
    furthest verified stage &
    Last evidenced stage, such as localization, attachment, placement
    estimate, release, or official reward. \\
    cost &
    Turns, Tool calls, model tokens, wall time, timeouts, and optional cost. \\
    violations &
    Unresolved observation/selection obligations, premature release, illegal
    numeric experience, and other contract violations. \\
    evidence index &
    Stable ID and content hash for rollout bundles, key frames, or video. \\
    \bottomrule
  \end{tabularx}
\end{table}

Table~\ref{tab:libero-results} reports suite aggregates. The four generated
tables below list, for all \LiberoTaskCount{} tasks, the number of successes
over \LiberoSeedCount{} seeds, mutually exclusive terminal summary, mean
resources, and content-addressed evidence ID. All values and macros derive
from one sanitized audit snapshot; development traces and obsolete
placeholders never enter the denominator.

For this fixed matrix, three simulator-TTL-contaminated cells lack complete
session results. Batch outcomes and the machine audit jointly fix their
identity, and they remain in each task's ten-seed denominator. Per-task
resource means use only complete episode records. The importer cannot rename
episode success as Pass@\(k\), and the publication gate re-renders every
formal value from the snapshot.

\ifdefstring{\LiberoResultStatus}{frozen}
  {
    \ifdefstring{\LiberoProtocolKind}{fixed-matrix}
      {
\subsection{Per-Task Fixed-Matrix Results}
\label{app:libero-fixed-results}

These tables are generated from completed batch \texttt{libero-fixed-40x10-20260730-r1} after receipt and episode-content-hash verification. Every task runs seeds 0--9 once. The metric is fixed-matrix episode success, not task-level @\(k\), and uses no qualification filter.
3 cells are infrastructure contaminated. They remain zeros in the preregistered 400-episode denominator but are omitted from per-episode means for turns, Tool calls, and tokens.

\begin{table}[htbp]
  \centering
  \scriptsize
  \renewcommand{\arraystretch}{1.10}
  \caption{\textbf{Spatial.}}
  \begin{tabularx}{\textwidth}{@{}lccccX@{}}
    \toprule
    \textbf{Task} & \textbf{Success} & \textbf{Rate} & \textbf{Resource \(n\)} & \textbf{Mean T/C} & \textbf{Primary failure} \\
    \midrule
    0 & 3/10 & 30.0\% & 10/10 & 37.7/37.3 & \texttt{episode\_timeout} \\
    1 & 0/10 & 0.0\% & 10/10 & 55.1/54.9 & \texttt{episode\_timeout} \\
    2 & 1/10 & 10.0\% & 10/10 & 27.0/27.0 & \texttt{episode\_timeout} \\
    3 & 1/10 & 10.0\% & 10/10 & 42.4/42.2 & \texttt{episode\_timeout} \\
    4 & 2/10 & 20.0\% & 10/10 & 41.4/41.2 & \texttt{episode\_timeout} \\
    5 & 0/10 & 0.0\% & 10/10 & 55.3/55.1 & \texttt{episode\_timeout} \\
    6 & 0/10 & 0.0\% & 10/10 & 42.2/41.9 & \texttt{episode\_timeout} \\
    7 & 0/10 & 0.0\% & 10/10 & 65.0/64.9 & \texttt{episode\_timeout} \\
    8 & 0/10 & 0.0\% & 10/10 & 50.9/50.5 & \texttt{episode\_timeout} \\
    9 & 1/10 & 10.0\% & 10/10 & 57.9/57.8 & \texttt{episode\_timeout} \\
    \bottomrule
  \end{tabularx}
\end{table}

\begin{table}[htbp]
  \centering
  \scriptsize
  \renewcommand{\arraystretch}{1.10}
  \caption{\textbf{Object.}}
  \begin{tabularx}{\textwidth}{@{}lccccX@{}}
    \toprule
    \textbf{Task} & \textbf{Success} & \textbf{Rate} & \textbf{Resource \(n\)} & \textbf{Mean T/C} & \textbf{Primary failure} \\
    \midrule
    0 & 0/10 & 0.0\% & 10/10 & 31.7/31.3 & \texttt{episode\_timeout} \\
    1 & 7/10 & 70.0\% & 10/10 & 36.1/36.1 & \texttt{episode\_timeout} \\
    2 & 2/10 & 20.0\% & 10/10 & 20.6/20.3 & \texttt{episode\_timeout} \\
    3 & 2/10 & 20.0\% & 10/10 & 40.2/40.1 & \texttt{episode\_timeout} \\
    4 & 4/10 & 40.0\% & 10/10 & 42.0/41.8 & \texttt{episode\_timeout} \\
    5 & 4/10 & 40.0\% & 10/10 & 41.9/41.8 & \texttt{episode\_timeout} \\
    6 & 3/10 & 30.0\% & 10/10 & 36.1/36.0 & \texttt{episode\_timeout} \\
    7 & 1/10 & 10.0\% & 10/10 & 33.3/33.2 & \texttt{episode\_timeout} \\
    8 & 0/10 & 0.0\% & 10/10 & 7.2/7.1 & \texttt{episode\_timeout} \\
    9 & 3/10 & 30.0\% & 10/10 & 23.4/23.2 & \texttt{episode\_timeout} \\
    \bottomrule
  \end{tabularx}
\end{table}

\begin{table}[htbp]
  \centering
  \scriptsize
  \renewcommand{\arraystretch}{1.10}
  \caption{\textbf{Goal.}}
  \begin{tabularx}{\textwidth}{@{}lccccX@{}}
    \toprule
    \textbf{Task} & \textbf{Success} & \textbf{Rate} & \textbf{Resource \(n\)} & \textbf{Mean T/C} & \textbf{Primary failure} \\
    \midrule
    0 & 0/10 & 0.0\% & 10/10 & 19.2/18.8 & \texttt{episode\_timeout} \\
    1 & 5/10 & 50.0\% & 10/10 & 61.9/61.9 & \texttt{max\_turns} \\
    2 & 0/10 & 0.0\% & 10/10 & 47.8/47.7 & \texttt{episode\_timeout} \\
    3 & 0/10 & 0.0\% & 10/10 & 37.9/37.7 & \texttt{episode\_timeout} \\
    4 & 9/10 & 90.0\% & 10/10 & 34.4/34.4 & \texttt{episode\_timeout} \\
    5 & 0/10 & 0.0\% & 10/10 & 57.1/57.1 & \texttt{episode\_timeout} \\
    6 & 5/10 & 50.0\% & 10/10 & 34.9/34.6 & \texttt{status\_report\_without\_reward} \\
    7 & 0/10 & 0.0\% & 10/10 & 7.2/7.2 & \texttt{episode\_timeout} \\
    8 & 2/10 & 20.0\% & 10/10 & 50.6/50.2 & \texttt{episode\_timeout} \\
    9 & 0/10 & 0.0\% & 10/10 & 16.8/16.3 & \texttt{episode\_timeout} \\
    \bottomrule
  \end{tabularx}
\end{table}

\begin{table}[htbp]
  \centering
  \scriptsize
  \renewcommand{\arraystretch}{1.10}
  \caption{\textbf{Long / LIBERO-10.}}
  \begin{tabularx}{\textwidth}{@{}lccccX@{}}
    \toprule
    \textbf{Task} & \textbf{Success} & \textbf{Rate} & \textbf{Resource \(n\)} & \textbf{Mean T/C} & \textbf{Primary failure} \\
    \midrule
    0 & 0/10 & 0.0\% & 10/10 & 101.7/101.3 & \texttt{max\_turns} \\
    1 & 0/10 & 0.0\% & 10/10 & 22.3/22.0 & \texttt{episode\_timeout} \\
    2 & 0/10 & 0.0\% & 10/10 & 30.5/29.8 & \texttt{unattended\_ask\_human} \\
    3 & 0/10 & 0.0\% & 10/10 & 49.1/48.7 & \texttt{episode\_timeout} \\
    4 & 0/10 & 0.0\% & 10/10 & 83.6/83.6 & \texttt{episode\_timeout} \\
    5 & 0/10 & 0.0\% & 10/10 & 78.3/78.0 & \texttt{episode\_timeout} \\
    6 & 0/10 & 0.0\% & 9/10 & 75.7/75.6 & \texttt{episode\_timeout} \\
    7 & 1/10 & 10.0\% & 10/10 & 105.3/105.2 & \texttt{episode\_timeout} \\
    8 & 0/10 & 0.0\% & 8/10 & 14.4/14.2 & \texttt{episode\_timeout} \\
    9 & 0/10 & 0.0\% & 10/10 & 17.5/16.5 & \texttt{unattended\_ask\_human} \\
    \bottomrule
  \end{tabularx}
\end{table}

Evidence ID \texttt{libero-fixed-40x10-20260730-r1:<suite>:task-<index>} links complete task text, trusted per-seed verdicts, resources, and source-record hashes to the release rollout index.
}
      {\input{sections/generated/libero_heldout_results}}
  }
  {}

Tables~\ref{tab:self-task-local-detail} and
\ref{tab:playbook-task-detail} provide per-task evidence for the
self-evolution comparisons. They preserve an important reporting property:
equal official reward does not erase differences in stage reachability or
resources, although those diagnostics do not establish capability equality or
improvement.

\subsection{Qualitative local success rollouts}

\begin{figure}[htbp]
  \centering
  \includegraphics[width=\textwidth]{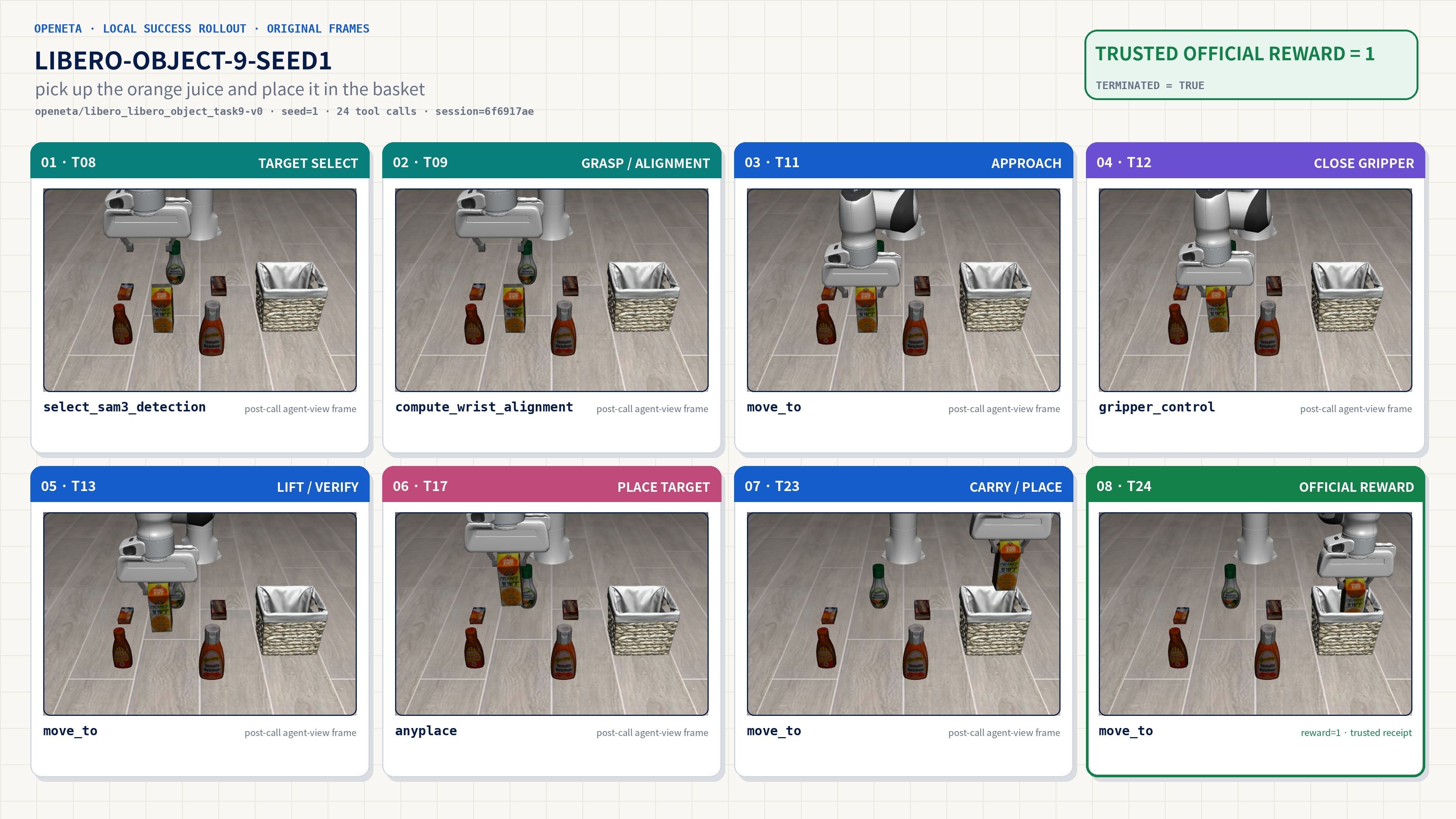}
  \caption{\textbf{Qualitative local Object-suite rollout.}
  The plate shows target selection, wrist alignment, approach, grasp,
  placement, and trusted reward as individually inspectable stages. It is
  excluded from the formal fixed-matrix denominator.}
  \label{fig:qualitative-libero-object}
\end{figure}

\begin{figure}[htbp]
  \centering
  \includegraphics[width=\textwidth]{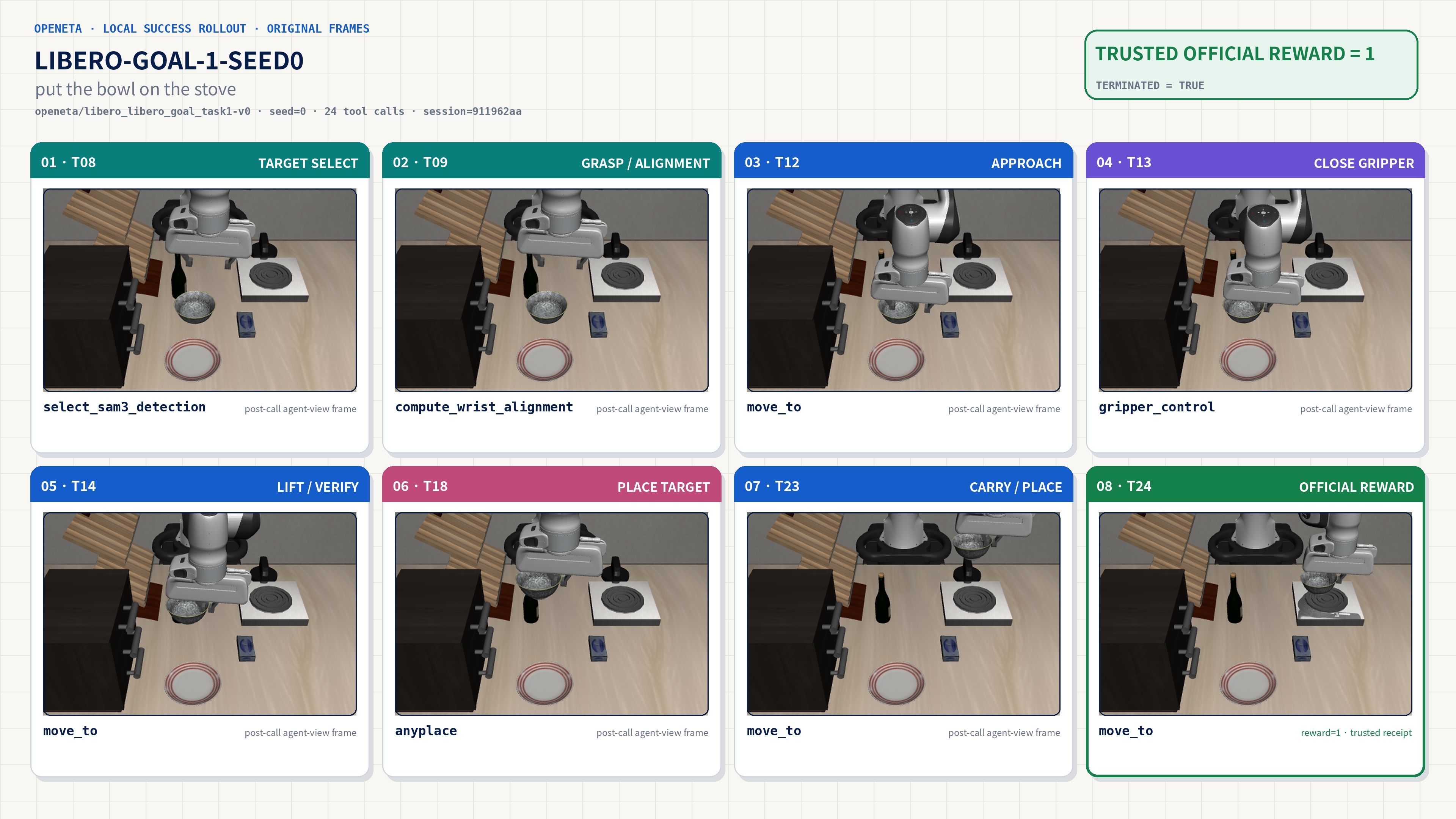}
  \caption{\textbf{Qualitative local Goal-suite rollout.}
  Fresh post-call frames make the grasp, lift, transport, placement, and
  official reward transitions visible. It is excluded from the formal
  fixed-matrix denominator.}
  \label{fig:qualitative-libero-goal}
\end{figure}

\begin{figure}[htbp]
  \centering
  \includegraphics[width=\textwidth]{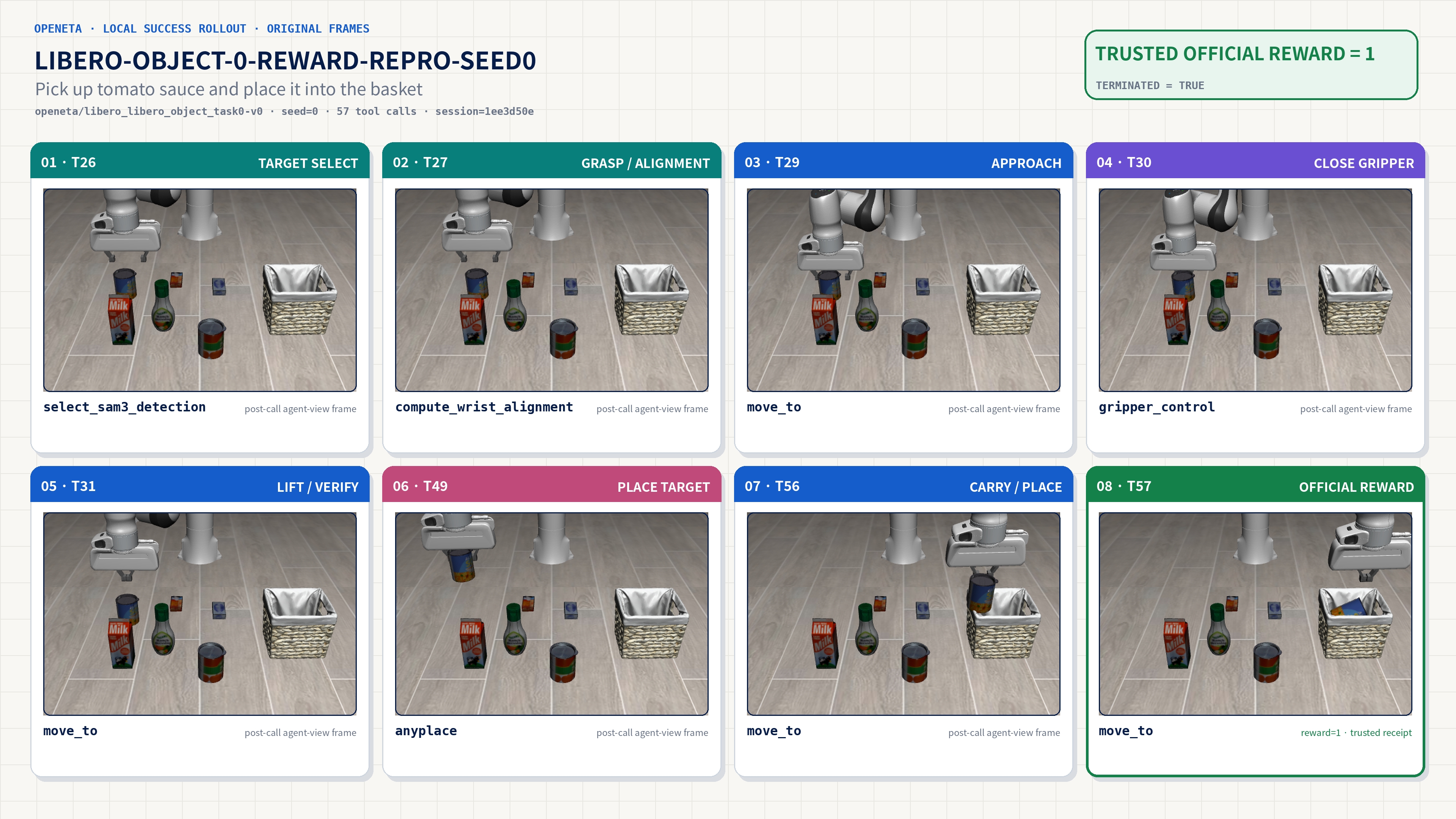}
  \caption{\textbf{Qualitative local reward-reproduction rollout.}
  This diagnostic plate visualizes a locally reproduced official reward and
  is not the frozen result for the similarly numbered Object-suite task. It
  is excluded from every formal aggregate and task-level metric.}
  \label{fig:qualitative-libero-reward-repro}
\end{figure}

The trajectory plates below demonstrate the report's visual trace format:
post-Tool observations, stage labels, Tool names, and the final trusted
environment reward remain visible in one artifact. They are local success
rollouts prepared for qualitative inspection. They are not indexed cells of
the frozen 400-episode matrix, are not used to recompute any table, and must
not be interpreted as additional benchmark successes.
Figures~\ref{fig:qualitative-libero-object},
\ref{fig:qualitative-libero-goal}, and
\ref{fig:qualitative-libero-reward-repro} instantiate this format for an
Object-suite task, a Goal-suite task, and a reward-reproduction diagnostic,
respectively.

\FloatBarrier

\section{Failure Taxonomy and Evidence Standard}
\label{app:failure-evidence}

OpenETA distinguishes call completion, physical action completion, and task
completion. A successful \tool{move_to} says that the backend completed one
atomic motion; attachment PASS says that current evidence supports continued
carriage; only trusted environment reward establishes task completion.

\begin{table}[htbp]
  \centering
  \scriptsize
  \renewcommand{\arraystretch}{1.15}
  \caption{\textbf{Recommended mutually exclusive primary episode failures.}}
  \label{tab:failure-taxonomy}
  \begin{tabularx}{\textwidth}{
    @{}p{0.24\textwidth}p{0.26\textwidth}X@{}}
    \toprule
    \textbf{Class} & \textbf{Decision boundary} & \textbf{Minimum evidence} \\
    \midrule
    \tool{infrastructure_invalid} &
    Environment, service, configuration, or logging prevents fair evaluation &
    Startup/health evidence, exception, and frozen-field drift; retain or
    exclude only under the preregistered denominator rule. \\
    \tool{target_localization_exhausted} &
    Target identity remains unconfirmed after allowed retrieval,
    localization, and segmentation &
    Latest observation, candidates/masks, and rejection reasons. \\
    \tool{grasp_proposal_or_safety} &
    No grasp candidate exists or all candidates fail geometry/safety gates &
    Candidate provenance, scores, frame transforms, and gate diagnostics. \\
    \tool{precontact_or_contact_motion} &
    Hover, approach, contact, close, or lift motion fails &
    Pre/post observations, action receipt, and controller diagnostics. \\
    \tool{attachment_fail_or_unknown} &
    Evidence shows the grasp failed or cannot confirm attachment &
    Probe type, threshold/diagnostic, and fresh observation; unknown is never
    PASS. \\
    \tool{placement_estimation} &
    No executable placement relation or pose is generated and verified &
    Receptacle selection, candidates, frames, and safety checks. \\
    \tool{premature_release_or_invariant} &
    Gripper opens before release prerequisites, or another obligation is
    violated &
    Obligation state and ordered accepted/blocked commands. \\
    \tool{post_release_no_reward} &
    Release and observation refresh complete without official success &
    Release receipt, fresh observation, and trusted environment receipt. \\
    \tool{resource_exhausted} &
    A valid run consumes its turns, calls, tokens, or wall-time budget &
    Terminal counters and unfinished stage. \\
    \tool{need_human_or_safe_stop} &
    Ambiguity or risk exceeds the autonomous boundary &
    Risk rationale, final safe state, and unresolved obligations. \\
    \bottomrule
  \end{tabularx}
\end{table}

\subsection{Minimum evidence chain for a case study}

A failure case should not be a single aftermath image. In temporal order it
should show: (1) task and initial observation; (2) target/candidate selection;
(3) last verified stage; (4) the decisive action, fresh post-action
observation, and trusted receipt; and (5) stop reason, remaining budget, and
unresolved obligations. Media are labeled with episode ID, time step, and
camera, while content hashes link the paper's key frames to the full bundle.

The primary class is mechanically selected from terminal evidence under a
fixed priority, one per episode. Secondary labels such as \tool{timeout},
\tool{segmentation_ambiguous}, or \tool{attachment_regression} may coexist.
This preserves an additive distribution without discarding cross-stage
symptoms. Insufficient evidence is \tool{unknown}, not an inferred root cause.

\FloatBarrier
\ifdefstring{\LiberoResultStatus}{frozen}
  {
    \ifdefstring{\LiberoProtocolKind}{fixed-matrix}
      {
\subsection{Representative Terminal Cases from the Formal LIBERO Batch}
\label{app:libero-formal-failure-cases}

Each row is selected mechanically rather than curated as a ``typical story.'' For every mutually exclusive terminal class, we choose the first failed episode ordered by preregistered suite, task index, and seed. Classes appear in decreasing batch frequency; Table~\ref{tab:libero-failure-taxonomy} gives complete counts.

{\scriptsize
\renewcommand{\arraystretch}{1.15}
\begin{longtable}{@{}p{0.19\textwidth}rp{0.13\textwidth}p{0.08\textwidth}p{0.17\textwidth}>{\RaggedRight\arraybackslash}p{0.25\textwidth}@{}}
  \caption{\textbf{Deterministic representative episode for each terminal class.} Resources are turns / Tool calls / wall-clock seconds.}\label{tab:libero-formal-failure-cases}\\
    \toprule
    \textbf{Terminal class} & \textbf{\(n\)} & \textbf{Suite} & \textbf{Task/seed} & \textbf{Resources} & \textbf{Evidence ID} \\
    \midrule
    \endfirsthead
    \toprule
    \textbf{Terminal class} & \textbf{\(n\)} & \textbf{Suite} & \textbf{Task/seed} & \textbf{Resources} & \textbf{Evidence ID} \\
    \midrule
    \endhead
    \texttt{episode\_timeout} & 215 & Spatial & 0/1 & 43/43/1801.5 & \path{libero-fixed-40x10-20260730-r1:libero_spatial:task-00:seed-1} \\
    \texttt{unattended\_ask\_human} & 66 & Spatial & 0/6 & 21/20/1382.1 & \path{libero-fixed-40x10-20260730-r1:libero_spatial:task-00:seed-6} \\
    \texttt{max\_turns} & 35 & Spatial & 1/4 & 100/100/554.6 & \path{libero-fixed-40x10-20260730-r1:libero_spatial:task-01:seed-4} \\
    \texttt{status\_report\_without\_reward} & 24 & Spatial & 0/0 & 33/32/1224.8 & \path{libero-fixed-40x10-20260730-r1:libero_spatial:task-00:seed-0} \\
    \texttt{simulator\_unknown\_handle} & 3 & Long / LIBERO-10 & 6/6 & --/--/1909.1 & \path{libero-fixed-40x10-20260730-r1:libero_10:task-06:seed-6} \\
    \texttt{remote\_episode\_terminated\_without\_reward} & 1 & Long / LIBERO-10 & 0/2 & 75/75/2735.7 & \path{libero-fixed-40x10-20260730-r1:libero_10:task-00:seed-2} \\
    \bottomrule
\end{longtable}
}

An Evidence ID identifies a task bundle and seed in the release rollout index; raw session IDs, local paths, service addresses, and trajectory payloads do not enter the paper. A terminal label says where execution stopped, not the unique physical or planning root cause. Infrastructure rows show ``--'' for turns and Tool calls because their batch outcomes lack complete resource fields. Root-cause claims still require pre/post observations, trusted receipts, and stage diagnostics.
}
      {}
  }
  {}
\clearpage

\subsection{Representative cases from frozen comparisons}

Table~\ref{tab:failure-case-playbook} uses a preregistered deterministic rule:
select the lexicographically first task and seed among all baseline-only
exact-task-playbook pairs. Both arms use task
\SelfPlaybookCaseTask{} and seed \SelfPlaybookCaseSeed{} in read-only mode;
only the candidate loads the exact-scope playbook.

\begin{table}[htbp]
  \centering
  \scriptsize
  \renewcommand{\arraystretch}{1.18}
  \caption{\textbf{Representative baseline-only exact-task-playbook pair.}
  Values come from the frozen snapshot; success means trusted reward only.}
  \label{tab:failure-case-playbook}
  \begin{tabularx}{\textwidth}{@{}p{0.22\textwidth}X X@{}}
    \toprule
    \textbf{Ordered evidence} & \textbf{Baseline} & \textbf{Candidate} \\
    \midrule
    1. Frozen condition &
    task \SelfPlaybookCaseTask, seed \SelfPlaybookCaseSeed; no playbook &
    task \SelfPlaybookCaseTask, seed \SelfPlaybookCaseSeed; exact-scope
    playbook loaded \\
    2. Trusted verdict &
    official success = \SelfPlaybookCaseBaselineSuccess &
    official success = \SelfPlaybookCaseCandidateSuccess \\
    3. Terminal state &
    environment completion (\tool{environment}) &
    episode wall-time exhaustion (\tool{episode_timeout}) \\
    4. Resources &
    \SelfPlaybookCaseBaselineTurns{} turns,
    \SelfPlaybookCaseBaselineCalls{} calls,
    \SelfPlaybookCaseBaselineSeconds\,s &
    \SelfPlaybookCaseCandidateTurns{} turns,
    \SelfPlaybookCaseCandidateCalls{} calls,
    \SelfPlaybookCaseCandidateSeconds\,s \\
    5. Mechanical verdict &
    successful control arm &
    \tool{resource_exhausted}; pair is \tool{baseline_only}; reject promotion \\
    \bottomrule
  \end{tabularx}
\end{table}

This pair establishes that, under frozen conditions, the candidate fails to
reproduce baseline reward and consumes more resources, so the promotion gate
must reject it. It does not prove that one playbook sentence is the unique
physical cause of timeout; that would require rule-trigger instrumentation or
a finer intervention.

The second case is the only valid contrastive-v2 replay. It separates the
common stage prefix, release obligation, official reward, and terminal state.

\begin{table}[htbp]
  \centering
  \scriptsize
  \renewcommand{\arraystretch}{1.18}
  \caption{\textbf{Invariant case for a contrastive stage-local strategy.}
  Neither arm receives reward; the candidate adds one premature gripper-open
  violation.}
  \label{tab:failure-case-contrastive-v2}
  \begin{tabularx}{\textwidth}{@{}p{0.22\textwidth}X X@{}}
    \toprule
    \textbf{Ordered evidence} & \textbf{Baseline} & \textbf{Candidate} \\
    \midrule
    1. Frozen condition &
    task \SelfStrategyTwoTask, seed \SelfStrategyTwoSeed; candidate hidden &
    task \SelfStrategyTwoTask, seed \SelfStrategyTwoSeed; reviewed candidate
    visible \\
    2. Shared prefix &
    segmentation $\rightarrow$ grasp estimation $\rightarrow$ contact
    $\rightarrow$ attachment PASS $\rightarrow$ placement estimate &
    same as baseline \\
    3. Release and reward &
    no valid placement release; official success =
    \SelfStrategyTwoBaselineSuccess &
    no valid placement release; official success =
    \SelfStrategyTwoCandidateSuccess \\
    4. Invariant evidence &
    \SelfStrategyTwoBaselineViolations{} violations &
    \SelfStrategyTwoCandidateViolations{}
    \path{open_before_attachment_failure_or_placement_release} \\
    5. Terminal/resources &
    \tool{episode_timeout};
    \SelfStrategyTwoBaselineTurns{} turns /
    \SelfStrategyTwoBaselineCalls{} calls /
    \SelfStrategyTwoBaselineSeconds\,s &
    \tool{status_report};
    \SelfStrategyTwoCandidateTurns{} turns /
    \SelfStrategyTwoCandidateCalls{} calls /
    \SelfStrategyTwoCandidateSeconds\,s \\
    6. Mechanical verdict &
    \tool{resource_exhausted} &
    \tool{premature_release_or_invariant}; replay fails, held-out not scheduled,
    promotion rejected \\
    \bottomrule
  \end{tabularx}
\end{table}

The valid pair has \SelfStrategyTwoExcludedInfrastructure{} infrastructure
exclusions, so it is eligible for task-level diagnosis. Operator
interruption, provider exhaustion, or shared-environment failure would instead
invalidate the entire pair. The evidence is sufficient to reject a candidate
that adds a violation without success; it does not establish that the
violation uniquely caused the absent reward.

\section{Supplementary Self-Evolution Results}
\label{app:self-evolution}

This appendix expands the paired studies in
Section~\ref{sec:self_revolution}. Success always means trusted official
reward; stage counts are diagnostic. A candidate that fails replay never
enters held-out evaluation, and an infrastructure-invalid pair is excluded.

\subsection{Task-local Skill/strategy}

\begin{table}[htbp]
  \centering
  \small
  \caption{\textbf{Per-task task-local results on
  \SelfTaskLocalSeeds{} held-out seeds.}}
  \label{tab:self-task-local-detail}
  \begin{tabular}{@{}lrrrrrr@{}}
    \toprule
    & \multicolumn{2}{c}{\textbf{Official success}} &
      \multicolumn{2}{c}{\textbf{Attachment PASS}} &
      \multicolumn{2}{c}{\textbf{Reached AnyPlace}} \\
    \cmidrule(lr){2-3}\cmidrule(lr){4-5}\cmidrule(l){6-7}
    \textbf{Spatial task} &
    \textbf{Base} & \textbf{Cand.} &
    \textbf{Base} & \textbf{Cand.} &
    \textbf{Base} & \textbf{Cand.} \\
    \midrule
    Task 1 &
    \SelfTaskOneOriginalSuccess/\SelfTaskLocalSeeds &
    \SelfTaskOneCandidateSuccess/\SelfTaskLocalSeeds &
    \SelfTaskOneOriginalAttachment & \SelfTaskOneCandidateAttachment &
    \SelfTaskOneOriginalAnyPlace & \SelfTaskOneCandidateAnyPlace \\
    Task 2 &
    \SelfTaskTwoOriginalSuccess/\SelfTaskLocalSeeds &
    \SelfTaskTwoCandidateSuccess/\SelfTaskLocalSeeds &
    \SelfTaskTwoOriginalAttachment & \SelfTaskTwoCandidateAttachment &
    \SelfTaskTwoOriginalAnyPlace & \SelfTaskTwoCandidateAnyPlace \\
    Task 4 &
    \SelfTaskFourOriginalSuccess/\SelfTaskLocalSeeds &
    \SelfTaskFourCandidateSuccess/\SelfTaskLocalSeeds &
    \SelfTaskFourOriginalAttachment & \SelfTaskFourCandidateAttachment &
    \SelfTaskFourOriginalAnyPlace & \SelfTaskFourCandidateAnyPlace \\
    \midrule
    Total &
    \SelfTaskLocalOriginalSuccess/\SelfTaskLocalEpisodesPerArm &
    \SelfTaskLocalCandidateSuccess/\SelfTaskLocalEpisodesPerArm &
    \SelfTaskLocalOriginalAttachment & \SelfTaskLocalCandidateAttachment &
    \SelfTaskLocalOriginalAnyPlace & \SelfTaskLocalCandidateAnyPlace \\
    \bottomrule
  \end{tabular}
\end{table}

Equal final reward does not mean no treatment effect: baseline/candidate arms
record
\SelfTaskLocalOriginalAttachment/\SelfTaskLocalCandidateAttachment{}
attachment passes and
\SelfTaskLocalOriginalAnyPlace/\SelfTaskLocalCandidateAnyPlace{} AnyPlace
reaches. These exploratory stage metrics have no preregistered inference test,
so they support only ``no observed success gain'' and preserve reduced upstream
reachability as a diagnostic, not a statistically significant degradation.

\subsection{Exact-task playbook}

\begin{table}[htbp]
  \centering
  \small
  \caption{\textbf{Per-task held-out pairs for the exact-task playbook.}}
  \label{tab:playbook-task-detail}
  \begin{tabular}{@{}lccc@{}}
    \toprule
    \textbf{Spatial task} & \textbf{Seeds} &
    \textbf{Baseline success} & \textbf{Playbook success} \\
    \midrule
    Task 1 & \SelfPlaybookSeeds &
    \SelfPlaybookTaskOneBaselineSuccess & \SelfPlaybookTaskOneCandidateSuccess \\
    Task 2 & \SelfPlaybookSeeds &
    \SelfPlaybookTaskTwoBaselineSuccess & \SelfPlaybookTaskTwoCandidateSuccess \\
    Task 4 & \SelfPlaybookSeeds &
    \SelfPlaybookTaskFourBaselineSuccess & \SelfPlaybookTaskFourCandidateSuccess \\
    \midrule
    Total & \SelfPlaybookEpisodesPerArm &
    \SelfPlaybookBaselineSuccess & \SelfPlaybookCandidateSuccess \\
    \bottomrule
  \end{tabular}
\end{table}

Baseline success is
\(\SelfPlaybookBaselineSuccess/\SelfPlaybookEpisodesPerArm
=\SelfPlaybookBaselineRatePercent\%\).
Playbook success is
\(\SelfPlaybookCandidateSuccess/\SelfPlaybookEpisodesPerArm
=\SelfPlaybookCandidateRatePercent\%\).
Among \SelfPlaybookEpisodesPerArm{} pairs, baseline-only =
\SelfPlaybookBaselineOnly, playbook-only = \SelfPlaybookCandidateOnly,
both-success = \SelfPlaybookBothSuccess, and both-fail =
\SelfPlaybookBothFail. The exact two-sided McNemar
\(p=\SelfPlaybookMcNemarP\) is insufficient for a significant degradation, but
the candidate fails the required ``objective gain without regression'' gate.

\begin{table}[htbp]
  \centering
  \scriptsize
  \renewcommand{\arraystretch}{1.12}
  \caption{\textbf{Stage reachability and resource diagnostics for the
  exact-task playbook.}}
  \label{tab:playbook-stage-cost}
  \begin{tabularx}{\textwidth}{@{}Xrr@{}}
    \toprule
    \textbf{Metric (each arm: \SelfPlaybookEpisodesPerArm{} episodes)} &
    \textbf{Baseline} & \textbf{Playbook} \\
    \midrule
    Reached grasp estimate & \SelfPlaybookBaselineGraspEstimate & \SelfPlaybookCandidateGraspEstimate \\
    Reached move & \SelfPlaybookBaselineMove & \SelfPlaybookCandidateMove \\
    Reached close & \SelfPlaybookBaselineClose & \SelfPlaybookCandidateClose \\
    Reached attachment assess & \SelfPlaybookBaselineAttachmentAssess & \SelfPlaybookCandidateAttachmentAssess \\
    Attachment PASS & \SelfPlaybookBaselineAttachmentPass & \SelfPlaybookCandidateAttachmentPass \\
    Reached AnyPlace & \SelfPlaybookBaselineAnyPlace & \SelfPlaybookCandidateAnyPlace \\
    Reached release & \SelfPlaybookBaselineRelease & \SelfPlaybookCandidateRelease \\
    Official reward & \SelfPlaybookBaselineReward & \SelfPlaybookCandidateReward \\
    \midrule
    Mean Tool calls & \SelfPlaybookBaselineMeanCalls & \SelfPlaybookCandidateMeanCalls \\
    Mean planner turns & \SelfPlaybookBaselineMeanTurns & \SelfPlaybookCandidateMeanTurns \\
    Mean wall time (s) & \SelfPlaybookBaselineMeanSeconds & \SelfPlaybookCandidateMeanSeconds \\
    Timeout episodes & \SelfPlaybookBaselineTimeouts & \SelfPlaybookCandidateTimeouts \\
    \bottomrule
  \end{tabularx}
\end{table}

Both versions produce a grasp estimate in every episode. They differ at later
stages. The baseline/playbook reach attachment assessment
\SelfPlaybookBaselineAttachmentAssess/\SelfPlaybookCandidateAttachmentAssess{}
times, pass the attachment check
\SelfPlaybookBaselineAttachmentPass/\SelfPlaybookCandidateAttachmentPass{}
times, and reach AnyPlace
\SelfPlaybookBaselineAnyPlace/\SelfPlaybookCandidateAnyPlace{} times. The
playbook also uses more turns and Tool calls and has more timeouts. These counts
show where the behaviors separate. They are not transition probabilities,
because a recovery path can skip a logged stage.

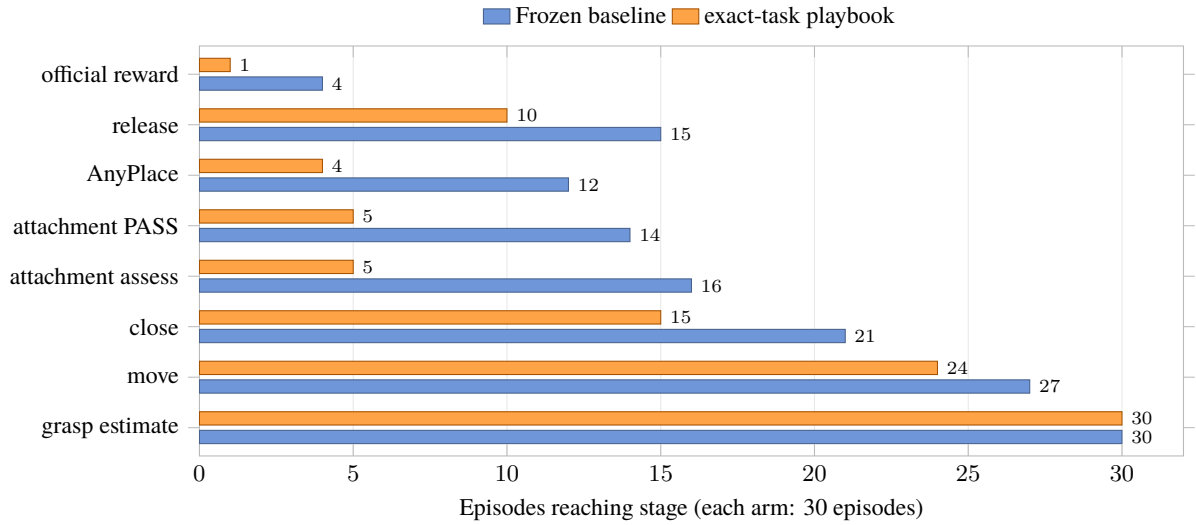
\begin{figure}[htbp]
  \centering
  \begin{tikzpicture}
    \begin{axis}[
      xbar,
      width=0.88\textwidth,
      height=7.0cm,
      xmin=0,
      xmax=32,
      xtick={0,5,10,15,20,25,30},
      xlabel={Episodes reaching stage (each arm:
        \SelfPlaybookEpisodesPerArm{} episodes)},
      symbolic y coords={
        official reward,
        release,
        AnyPlace,
        attachment PASS,
        attachment assess,
        close,
        move,
        grasp estimate
      },
      ytick=data,
      y dir=reverse,
      bar width=5pt,
      enlarge y limits=0.08,
      nodes near coords,
      point meta=x,
      every node near coord/.append style={font=\scriptsize},
      legend style={at={(0.5,1.02)},anchor=south,legend columns=2,draw=none,font=\small},
      legend image code/.code={\draw[#1] (0cm,-0.1cm) rectangle (0.35cm,0.1cm);},
      tick label style={font=\small},
      label style={font=\small},
      axis line style={draw=gray!55},
      tick style={draw=gray!55},
      xmajorgrids=true,
      grid style={gray!18}
    ]
      \addplot[fill=MossBlue!85!black,draw=MossBlue!55!black] coordinates {
        (\SelfPlaybookBaselineGraspEstimate,{grasp estimate})
        (\SelfPlaybookBaselineMove,{move})
        (\SelfPlaybookBaselineClose,{close})
        (\SelfPlaybookBaselineAttachmentAssess,{attachment assess})
        (\SelfPlaybookBaselineAttachmentPass,{attachment PASS})
        (\SelfPlaybookBaselineAnyPlace,{AnyPlace})
        (\SelfPlaybookBaselineRelease,{release})
        (\SelfPlaybookBaselineReward,{official reward})
      };
      \addplot[fill=orange!72,draw=orange!65!black] coordinates {
        (\SelfPlaybookCandidateGraspEstimate,{grasp estimate})
        (\SelfPlaybookCandidateMove,{move})
        (\SelfPlaybookCandidateClose,{close})
        (\SelfPlaybookCandidateAttachmentAssess,{attachment assess})
        (\SelfPlaybookCandidateAttachmentPass,{attachment PASS})
        (\SelfPlaybookCandidateAnyPlace,{AnyPlace})
        (\SelfPlaybookCandidateRelease,{release})
        (\SelfPlaybookCandidateReward,{official reward})
      };
      \legend{Frozen baseline,exact-task playbook}
    \end{axis}
  \end{tikzpicture}
  \caption{\textbf{Stage reach for the exact-task playbook.}
  Each arm contains \SelfPlaybookEpisodesPerArm{} paired episodes. The chart
  shows where the baseline and playbook begin to differ. Stage events are not
  a strictly nested funnel because recovery paths can skip events.}
  \label{fig:self-playbook-stage-profile}
\end{figure}

\subsection{Stage-local task-strategy}

\begin{table}[htbp]
  \centering
  \scriptsize
  \renewcommand{\arraystretch}{1.15}
  \caption{\textbf{Replay-gate results for stage-local task-strategy candidates.}}
  \label{tab:strategy-replay-detail}
  \begin{tabularx}{\textwidth}{
    @{}p{0.13\textwidth}p{0.19\textwidth}p{0.20\textwidth}X@{}}
    \toprule
    \textbf{Version} & \textbf{Valid comparison} &
    \textbf{Resources/outcome} & \textbf{Gate verdict} \\
    \midrule
    v1 &
    Tasks 1/2/4, seed 0, \SelfStrategyOnePairCount{} pairs &
    Both arms
    \SelfStrategyOneBaselineSuccess/\SelfStrategyOnePairCount{};
    baseline/candidate mean turns
    \SelfStrategyOneBaselineMeanTurns/\SelfStrategyOneCandidateMeanTurns{},
    mean time
    \SelfStrategyOneBaselineMeanSeconds/\SelfStrategyOneCandidateMeanSeconds{} s &
    Reward not reproduced and candidate slower; held-out not scheduled. \\
    v2 &
    Task 2, seed 0, \SelfStrategyTwoPairCount{} valid pair &
    Both arms
    \SelfStrategyTwoBaselineSuccess/\SelfStrategyTwoPairCount{};
    baseline \SelfStrategyTwoBaselineTurns{} turns /
    \SelfStrategyTwoBaselineCalls{} calls /
    \SelfStrategyTwoBaselineTokens{} tokens /
    \SelfStrategyTwoBaselineSeconds{} s; candidate
    \SelfStrategyTwoCandidateTurns{} /
    \SelfStrategyTwoCandidateCalls{} /
    \SelfStrategyTwoCandidateTokens{} /
    \SelfStrategyTwoCandidateSeconds{} s &
    No change to the key failure and
    \SelfStrategyTwoCandidateViolations{} new premature-open violation;
    reject. \\
    \bottomrule
  \end{tabularx}
\end{table}

A second Task-1 diagnostic pair in v2 fails batch-validity conditions and is
excluded. Experimental validity determines whether a comparison is
interpretable; the capability gate then decides whether a valid candidate
deserves promotion. All candidates remain isolated and none became a shared
capability.

\subsection{Validity boundaries}

\begin{itemize}[leftmargin=*]
  \item \textbf{Adaptive sequence.} Later representations and gates follow
  earlier diagnoses; the five studies are not independent replications under
  one preregistration. The McNemar value describes only the exact-task
  comparison.
  \item \textbf{Coverage.} Formal pairs concentrate on LIBERO Spatial Tasks
  1/2/4, with still smaller stage-local replay. Results do not generalize to
  all LIBERO, LIBERO-Pro, or real robots.
  \item \textbf{Residual randomness.} Matched task, seed, and budget control
  initial environment state, but remote model sampling, perception, and
  contact remain stochastic. Zero success in a small sample does not establish
  equivalence.
  \item \textbf{Treatment integrity.} Loading experience does not prove that a
  key rule triggered or was followed. V2 records adherence and violations;
  earlier studies infer them from Tool sequences and events.
  \item \textbf{Selection and credit.} A playbook from one success may encode
  incidental perception or contact. Non-nested stage events localize
  divergence but do not prove causality.
\end{itemize}

The strongest supported statement is that no candidate satisfies the declared
promotion criteria, not that textual experience can never improve a physical
task. A confirmatory study should freeze tasks, seeds, primary metric,
treatment-integrity checks, and thresholds before candidate generation, then
evaluate once on seeds excluded from candidate formation.

\subsection{Frozen settings and identity limits}

\begin{table}[htbp]
  \centering
  \small
  \renewcommand{\arraystretch}{1.12}
  \caption{\textbf{Common settings auditable from self-evolution manifests.}}
  \label{tab:self-evolution-reproducibility}
  \begin{tabularx}{\textwidth}{@{}p{0.25\textwidth}p{0.29\textwidth}X@{}}
    \toprule
    \textbf{Setting} & \textbf{Frozen value} & \textbf{Evidence boundary} \\
    \midrule
    Benchmark & LIBERO Spatial &
    Every episode binds task index, seed, and normalized task-text SHA-256. \\
    Common budget &
    At most \SelfEvolutionMaxTurns{} turns,
    \SelfEvolutionMaxToolCalls{} Tool calls,
    \SelfEvolutionMaxTokens{} tokens, and
    \SelfEvolutionTimeoutSeconds{} s &
    Every manifest entry in the six source batches carries these values. \\
    Success and task identity &
    Official environment reward; simulator-assigned task &
    Both conditions are required per episode; stages do not replace success. \\
    Paired evaluation &
    Same task, seed, and budget &
    Task-local, playbook, v1, and v2 are read-only pairs; online rounds are
    adaptive exploration. \\
    Primary planner identity &
    \SelfEvolutionPlannerIdentityStatus &
    Provider/model is absent from historical run manifests, preventing exact
    planner-version attribution. \\
    V2 author/reviewer &
    \SelfStrategyTwoAuthorProvider{} /
    \texttt{\SelfStrategyTwoAuthorModel}; two isolated contexts &
    \SelfStrategyTwoAcceptedCandidates{} candidates accepted. Author and
    reviewer both use \texttt{\SelfStrategyTwoReviewerModel}; this is
    context isolation, not cross-model review. \\
    \bottomrule
  \end{tabularx}
\end{table}

\subsection{Evidence index}

\begin{table}[htbp]
  \centering
  \small
  \caption{\textbf{Stable experiment identifiers for quantitative claims.}}
  \label{tab:self-evolution-evidence-index}
  \begin{tabularx}{\textwidth}{
    @{}>{\raggedright\arraybackslash\ttfamily\footnotesize}p{0.43\textwidth}X@{}}
    \toprule
    {\normalfont\bfseries Experiment ID} & \textbf{Evidence} \\
    \midrule
    \SelfOnlineEvidenceOne &
    First three-round online study:
    \(\SelfOnlineOneRoundOne\!\rightarrow\!\SelfOnlineOneRoundTwo
    \!\rightarrow\!\SelfOnlineOneRoundThree\). \\
    \SelfOnlineEvidenceTwo &
    Second three-round online study:
    \(\SelfOnlineTwoRoundOne\!\rightarrow\!\SelfOnlineTwoRoundTwo
    \!\rightarrow\!\SelfOnlineTwoRoundThree\). \\
    \SelfTaskLocalEvidence & Task-local paired outcomes and stages. \\
    \SelfPlaybookEvidence &
    \SelfPlaybookEpisodesPerArm{} exact-task held-out pairs. \\
    \SelfStrategyOneEvidence & Valid seed-0 stage-local v1 replay. \\
    \SelfStrategyTwoEvidence &
    Valid Task-2 contrastive-v2 replay; invalid batch excluded. \\
    \bottomrule
  \end{tabularx}
\end{table}

These IDs point to frozen summaries and rollout manifests, not local absolute
paths. The sanitized machine-readable snapshot is
\path{results/self_evolution_summary.json}. The freezing script verifies code
commit, dirty-diff hash, summary/analysis/gate hashes, batch validity, cleanup,
and promotion verdict, then deterministically generates both the snapshot and
\path{sections/generated/self_evolution_result_values.tex}. The publication
gate compares every macro and generated byte. With the six source directories,
\path{make verify-self-evolution-source OPENETA_MEMORY=<memory-root>} rebuilds
the source-to-snapshot-to-paper chain without overwriting repository files.

\section{Core Execution and Trajectory Protocol}
\label{app:core-protocol}

This appendix fixes the minimum protocol surface used by the paper. Tool
parameters may evolve, but no version may bypass the command types, trusted
receipts, or post-action observation invariant below. A release manifest states
the implemented schema versions and hashes.

\subsection{Commands and normalized results}

\begin{table}[htbp]
  \centering
  \small
  \renewcommand{\arraystretch}{1.15}
  \caption{\textbf{Stable top-level contract between planner and host.}}
  \label{tab:core-command-contract}
  \begin{tabularx}{\textwidth}{@{}p{0.24\textwidth}X@{}}
    \toprule
    \textbf{Object} & \textbf{Stable semantics} \\
    \midrule
    \tool{AgentCommand} &
    Schema \tool{openeta.agent_command.v1}; top-level \tool{kind} is
    \tool{tool_call} or \tool{response}. A Tool call names a host-registered
    Tool with structured arguments; a response has no physical side effect. \\
    \tool{ToolResult} &
    Schema \tool{openeta.tool_result.v1}; common fields include \tool{tool},
    \tool{category}, \tool{effect}, \tool{result_type}, and \tool{success}.
    Payload fields separately store \tool{parameters}, \tool{outputs},
    \tool{artifacts}, \tool{state_delta}, \tool{diagnostics}, and
    \tool{requires_observation_after_call}. \\
    Tool binding &
    The host owns Tool names, parameter schemas, handlers, backends, and side
    effects. A Skill or model output can select registered capabilities but
    cannot masquerade as a new atomic action. \\
    \bottomrule
  \end{tabularx}
\end{table}

\subsection{Trusted environment receipts}

Official reward and termination require a host-attested
\tool{openeta.environment_receipt.v1}. Runtime validation matches authority,
execution ID, session ID, and environment handle to the current call. Ordinary
Tool output or model text cannot mint official reward. Timeout means that a
side effect may have occurred; recovery queries backend state or observes
again rather than blindly replaying the mutation.

\subsection{Runtime obligations}

\begin{table}[htbp]
  \centering
  \small
  \renewcommand{\arraystretch}{1.15}
  \caption{\textbf{Obligations that constrain physical execution.}}
  \label{tab:runtime-obligations}
  \begin{tabularx}{\textwidth}{
    @{}p{0.23\textwidth}p{0.30\textwidth}X@{}}
    \toprule
    \textbf{Obligation} & \textbf{Trigger} &
    \textbf{Discharge condition and blocked scope} \\
    \midrule
    Fresh observation &
    A world-changing result lacks a sufficiently fresh snapshot &
    Trusted observation must update state before another state-dependent
    physical decision; stop safely after three failed refreshes. \\
    Target selection &
    Segmentation yields multiple candidates or identity is ambiguous &
    Record an explicit candidate and rationale; block grasp/contact until
    resolved. \\
    Grasp candidate &
    The planner switches or rejects a grasp proposal &
    Retain provenance, transform, score, and checks so candidate changes remain
    traceable. \\
    Release prerequisite &
    The object is believed attached and placement begins &
    Open the gripper only after placement relation, motion, and state evidence
    satisfy the contract; otherwise block and record a violation. \\
    \bottomrule
  \end{tabularx}
\end{table}

\subsection{Session state and release trajectories}

\path{trace.jsonl} stores audit events, \path{conversation.jsonl} supports
history recovery, \path{working/} stores mutable state, and \path{rollout/}
stores the immutable release bundle. Its \path{manifest.json} fixes
configuration. Files \path{model_calls.jsonl}, \path{tool_calls.jsonl},
\path{transitions.jsonl}, \path{episodes.jsonl}, and
\path{artifacts.jsonl} store calls, state changes, summaries, and
content-addressed artifacts. Training and paper statistics consume validated
rollout bundles, not console logs.

\subsection{Release-grade system-contract snapshot}

\path{scripts/freeze_system_contracts.py} generates
\path{results/system_contracts.json} from a clean OpenETA release checkout.
The snapshot freezes the Git commit; hashes of four critical source files; Tool
names, categories, side effects, fresh-observation requirements, and parameter
schemas; command, result, receipt, and rollout schema versions; root dependency
files; seven perception/grasp requirement files; and the UniDepth deployment
Dockerfile.

The freezer requires exactly 44 registered Tools and a clean checkout.
Development branches can create preview snapshots, which cannot pass the
publication gate. This report's release snapshot comes from clean
\texttt{main@135a7edc7e60}: 16 read-only, 12 planning, 9 bookkeeping, and 7
world-mutating Tools. Adding, removing, or reclassifying a Tool, or changing a
schema, explicitly invalidates the paper check. Dependency hashes establish
what the release source declares, not what was installed on the evaluation
machine; each experimental manifest records the latter independently.

\section{Sim2Real Disclosure and Safety Checklist}
\label{app:sim2real-safety}

Interface presence, primitive-level hardware validation, and complete
real-robot task success are distinct evidence levels. We use the terms
\emph{interface integration}, \emph{primitive validation}, and \emph{task
validation} separately; simulation success raises none of them automatically.

\subsection{Frozen interface-level evidence}

The current report is frozen at \textbf{interface integration}. The
machine-readable \path{results/real_robot_interface.json} snapshot comes from
clean release branch \texttt{main@135a7edc7e60}. It records six camera-driver
keys, two arm-driver keys, six lifecycle entry points, and five control entry
points, and hashes adapter, registry, MCP, driver, calibration, configuration,
and test source. Device IPs, serial numbers, and calibration matrices are not
exported.

The snapshot verifies source paths for cross-process exclusion, handle/session
ownership, reset-before-control, post-action observation, idempotent close, and
per-command translation/rotation limits; out-of-bounds requests are rejected,
not silently clipped. Fixed-base control remains an explicit stub. These are
source contracts, not evidence of SDK installation, device reachability,
current calibration, on-hardware primitive execution, collision safety, or
autonomous task success.

\subsection{Real-robot execution status}

\ifdefstring{\RealRobotResultStatus}{frozen}
  {
    \input{sections/generated/real_robot_setup}
    \input{sections/generated/real_robot_summary}
    \input{sections/generated/real_robot_trials}
  }
  {
    \begin{table}[htbp]
      \centering
      \small
      \caption{\textbf{Current real-robot evidence status.}}
      \label{tab:real-robot-result-status}
      \begin{tabularx}{\textwidth}{@{}p{0.25\textwidth}X@{}}
        \toprule
        \textbf{Item} & \textbf{Status} \\
        \midrule
        Formal batch & Pending; no development value is promoted into the
        formal result macros. \\
        Current citable level & Interface integration. \\
        Retained demonstrations &
        One complete sponge-to-tray recording; one bell-pepper recording with
        grasp and transport visible but no supported final in-basket relation.
        Neither recording establishes a denominator or success rate. \\
        Primitive/task results &
        No formal batch has passed completion-receipt, evidence-hash,
        intervention, and safety-stop audit. \\
        Formal import products &
        Sanitized JSON, result macros, reproducibility/safety settings,
        per-target table, and per-trial evidence index. \\
        \bottomrule
      \end{tabularx}
    \end{table}
  }

\subsection{Demonstration Setup, Observations, and Evidence Boundary}

Development testing used a RealSense D435i for the wrist view and a
supplementary third-person view, and a lidar-equipped RealSense L515 for the
main third-person view. Relative to simulator depth, missing and unreliable
depth regions propagated through depth enhancement into pose estimation. The
resulting target poses were often too low-quality for consistent grasping.
Separately, lateral grasp commands sometimes triggered acceleration-limit
protective stops. Controller PD parameters or trajectory shaping are plausible
contributors, but controlled replay and controller telemetry are required
before assigning cause.

The publication manifest cites a machine-readable record containing the two
video hashes and their visual-verdict boundaries. It deliberately remains
outside the formal result macros. Promotion requires a trial ledger with trial
order, outcome, terminal reason, intervention and safety-stop fields;
continuous video or trial-aligned clips; rollout and verdict hashes; the
executed clean commit and interface-contract hash; calibration statistics;
safety-limit and emergency-stop records; and a frozen protocol. Selected
recordings may illustrate behavior, but they cannot establish a denominator or
failure rate.

For public release, the project page should present the two videos as
\emph{representative development demonstrations}, link the evidence record,
and expose content hashes. Selected frames may be used in the paper after the
videos are frozen; each frame should identify its task and stage, and the
caption must retain the qualitative-demonstration qualifier.

\subsection{Automated audit of a completed batch}

A formal batch provides \path{run_manifest.json}, \path{trials.json},
\path{summary.json}, and \path{COMPLETED.json}. The importer verifies the
receipt and clean-code hashes. It also checks the interface contract, sanitized
hardware and firmware summaries, calibration error, emergency-stop test,
on-site supervision, preregistration, common budgets, and
video/rollout/verdict evidence for every trial. Public
snapshots exclude network addresses, serial numbers, local paths, handles,
person names, and credentials.

Autonomous success requires a valid trial, official-success termination, no
physical intervention, and no safety stop. Interventions and safety stops
remain autonomous failures. Preregistered infrastructure-invalid trials are
excluded from the rate denominator but retain invalidity evidence and counts.
The importer derives target summaries, trial indices, and
hardware, calibration, safety, protocol, planner, and budget settings from one
audited snapshot.

\subsection{Minimum disclosure for a physical batch}

\begin{table}[htbp]
  \centering
  \scriptsize
  \renewcommand{\arraystretch}{1.16}
  \caption{\textbf{Minimum archived or public evidence for each real-robot batch.}}
  \label{tab:sim2real-checklist}
  \begin{tabularx}{\textwidth}{
    @{}p{0.23\textwidth}p{0.30\textwidth}X@{}}
    \toprule
    \textbf{Item} & \textbf{Frozen information} & \textbf{Minimum evidence} \\
    \midrule
    Hardware/backend &
    Robot, gripper, camera, controller, firmware, and adapter versions &
    Device manifest, health checks, and backend hashes. \\
    Frames/calibration &
    Base, end-effector, camera, and World frames; intrinsic/extrinsic profile &
    Calibration date, errors, validation points, and valid distance. \\
    Timing &
    Timestamp sources and synchronization for image, depth, state, and receipt &
    Maximum staleness and observation-to-state alignment checks. \\
    Depth/geometry &
    Units, invalid-value handling, enhancement, and valid domain &
    Raw/enhanced depth references, scale checks, and boundary cases. \\
    Motion limits &
    Position, orientation, speed, acceleration, jerk, force/torque, and payload &
    Configuration plus tests showing out-of-range commands are blocked. \\
    Collision/contact &
    Self/environment collision, approach/contact modes, and recovery &
    Preview/check output, low-speed contact tests, and failure stops. \\
    Emergency stop/supervision &
    Hardware/software stop, operator position, takeover, and reset &
    Pre-batch stop test and supervision record. \\
    Request semantics &
    Command ID, timeout, cancellation, retry, and idempotency &
    Post-timeout state query and no unaudited duplicate motion. \\
    Post-action observation &
    Fresh snapshot return or explicit observation obligation &
    Time-aligned pre/post observations and discharge record. \\
    Resource ownership &
    Exclusive lease, session isolation, and abnormal release &
    Concurrent-request rejection and crash-safe reclamation. \\
    Task/evidence &
    Task, setup, trial, perturbation, budget, and success rule &
    Continuous video, structured rollout, official verdict, and failure class. \\
    Promotion boundary &
    Canary Tool, composed flow, full task, and held-out scene order &
    Independent pass without new safety violation at each level. \\
    \bottomrule
  \end{tabularx}
\end{table}

Every intervention is reported with its time. Risk-motivated takeover remains a
safety event and cannot be edited into an autonomous success. Human-shared
workspaces, unknown objects, and higher-speed studies require separate risk
assessment. This checklist is a paper-disclosure minimum, not a substitute for
manufacturer requirements, institutional safety procedures, or on-site
operator judgment.

\end{document}